%% file: main.tex
\documentclass[twoside,11pt]{article}

\usepackage{jair}
\usepackage{lmodern}
\usepackage{theapa}
\usepackage{rawfonts}
\usepackage{rotating}
\usepackage{pdflscape} 
\usepackage{booktabs} 
\usepackage{textcomp}
\usepackage[utf8]{inputenc} 
\usepackage[T1]{fontenc}  
\usepackage{url}  
\usepackage{amsfonts}  
\usepackage{nicefrac}  
\usepackage{microtype} 
\usepackage{multirow,array}
\usepackage{caption} 
\usepackage{longtable}
\usepackage{lipsum} 
\usepackage{amsmath}
\usepackage{amssymb}
\usepackage[title]{appendix}
\usepackage{float}
\usepackage{cleveref}

\usepackage{blindtext}
\usepackage{enumitem}
\usepackage{xcolor}
\usepackage{subfigure}
\usepackage{csvsimple}
\input{inputs/defs.tex}

\usepackage[export]{adjustbox}

\jairheading{73}{2022}{1025-1091}{11/2021}{03/2022}
\ShortHeadings{Predicting Decisions in Language Based Persuasion Games}
{Apel, Erev, Reichart \& Tennenholtz}

\firstpageno{1025}

\begin{document}

\title{Predicting Decisions in Language Based Persuasion Games}

\author{\name Reut Apel \email reutapel@campus.technion.ac.il \\
   \name Ido Erev \email erevido@gmail.com \\
    \name Roi Reichart \email roiri@ie.technion.ac.il \\
   \name Moshe Tennenholtz \email moshet@ie.technion.ac.il \\
   \addr Faculty of Industrial Engineering and Management\\
   Technion—Israel Institute of Technology, Israel}


\maketitle

\begin{abstract}

Sender-receiver interactions, and specifically persuasion games, are widely researched in economic modeling and artificial intelligence, and serve as a solid foundation for powerful applications. However, in the classic persuasion games setting, the messages sent from the expert to the decision-maker are abstract or well-structured application-specific signals rather than natural (human) language messages, although natural language is a very common communication signal in real-world persuasion setups. This paper addresses the use of natural language in persuasion games, exploring its impact on the decisions made by the players and aiming to construct effective models for the prediction of these decisions.

For this purpose, we conduct an online repeated interaction experiment. At each trial of the interaction, an informed expert aims to sell an uninformed decision-maker a vacation in a hotel, by sending her a review that describes the hotel. While the expert is exposed to several scored reviews, the decision-maker observes only the single review sent by the expert, and her payoff in case she chooses to take the hotel is a random draw from the review score distribution available to the expert only. The expert's payoff, in turn, depends on the number of times the decision-maker chooses the hotel. We also compare the behavioral patterns in this experiment to the equivalent patterns in similar experiments where the communication is based on the numerical values of the reviews rather than the reviews' text, and observe substantial differences which can be explained through an equilibrium analysis of the game.

We consider a number of modeling approaches for our verbal communication setup, differing from each other in the model type (deep neural network (DNN) vs. linear classifier), the type of features used by the model (textual, behavioral or both) and the source of the textual features (DNN-based vs. hand-crafted). Our results demonstrate that given a prefix of the interaction sequence, our models can predict the future decisions of the decision-maker, particularly when a sequential modeling approach and hand-crafted textual features are applied. Further analysis of the hand-crafted textual features allows us to make initial observations about the aspects of text that drive decision making in our setup. \footnote{Our code and data are available at: \url{https://github.com/reutapel/Predicting-Decisions-in-Language-Based-Persuasion-Games}}

\end{abstract}

\section{Introduction}
\label{Introduction}
The modeling and analysis of sender-receiver interactions are central to both economic modeling and Artificial Intelligence (AI). Indeed, the 2001 Nobel prize in economics was presented to Akerlof, Spence, and Stiglitz, for their pioneering research lines, showing how the signaling of information can alter strategic interactions (see \cite{Spence73}). Of particular interest is the study of cheap talk (\cite{crawford_sobel_1982}), and Bayesian persuasion (\cite{kamenica_gentzkow_2009}, following \cite{aumann_maschler_stearns_1995}), dealing with different levels of commitment power of the sender. The settings of personalized advertising and targeted recommendation systems \citep{shapiro_varian_2008,emek2014signaling}, where different services are offered to potential customers, are forms of strategic sender-receiver interactions in the spirit of these stylized economic models \citep{arieli_babichenko_2019}. In AI, the whole agenda of chat-bots design is targeted at sender-receiver interactions, emphasizing the use of language (\cite{jurafsky_martin_2018}).

Most economic models of sender-receiver interactions are game-theoretic ones. In such a setting, both sender and receiver are strategic players with their own private utilities. They, however, possess asymmetric information: Typically, the sender has more information than the receiver about the state of nature. Of particular interest is the persuasion game setting, where the sender's objective is to persuade the receiver to select some course of action among a set of possible actions. The receiver, in turn, has different payoffs for the different actions. While the receiver's payoff depends on the state of nature, he has a higher level of uncertainty about the state of nature than the sender has. The study of this fundamental setting has received a significant amount of attention in recent years and serves as a solid foundation for powerful applications (\cite{emek2014signaling}, \cite{bahar2016economic}). In what follows, we refer to the sender as an \textit{expert} and to the receiver as a \textit{decision-maker}.

In this work, we adopt the above framework of persuasion games and consider a setting of repeated expert-decision-maker games. Hence, while there is no notion of commitment by the expert in our setting, there are definitely possible reputation effects (\cite{kreps_milgrom_roberts_wilson_1982}). Bridging this foundational setting with language-oriented AI, our work introduces for the first time the use of natural language into these stylized persuasion games. Namely, while in the classical setting, the expert's messages to the decision-maker are abstract or well-structured application-specific signals (as nicely implemented also in AI settings: \cite{azaria2012strategic}, \cite{azaria2011giving}), in our setting these messages are expressed in natural language.

To be more precise, consider a setting where a decision-maker has to select among a safe constant-payoff action and a risky action with the payoff determined by some probability distribution unknown to the decision-maker. The expert aim is for the decision-maker to select the risky action. He can do it by communicating messages, where each message is associated with a different payoff in support of the distribution. The interaction is repeated, where at each interaction, a different distribution is selected, and different messages are associated with the different payoffs. The messages and their relationship to payoffs are grounded in some real-world events, e.g., messages associated with corresponding numeric grades of hotel reviews.
The main questions we ask: Given some game's history in the first K trials of interaction, can we predict behavior in the game's subsequent trials? What is the best way to come up with such a prediction?

The approach we have taken to tackle the above challenge is as follows. We created a data set that was collected using an online experiment. In the experiment, two participants are randomly and anonymously paired, and each of them was randomly selected to one of two roles: Decision-maker and expert. They then play a ten trial game together, where at each trial the expert is asked to select one of seven hotel reviews presented to her alongside their related scores. The chosen review was then presented to the decision-maker without its numerical score, and the decision-maker was asked to choose between the risky hotel and the safe stay home options. The expert benefits from hotel choice by the decision-maker, while the decision-maker's payoff in the hotel choice case was determined by the score distribution presented only to the expert.

We have also performed a series of additional experiments that aim to explore whether the verbal communication between the expert and the decision maker changes their behavior. Our analysis answers this question positively, and, particularly, demonstrates that verbal communication increases the descriptive value of a particular equilibrium that considers a simplified version of the game that we are able to theoretically analyse (see details below, as well as in Section \ref{sec:verbal}). These observations naturally increase our interest in understanding the impact of verbal communication on decision making in our setup.

Given our data and the above observations, we are interested in predicting the decision-makers' decisions, which determine the experts' payoffs. Notably, we define two sets of research questions. The first set is about the nature of the prediction task: (1) Given the history of the first $pr$ trials, can we predict the decision made by the decision-maker in each ensuing trial? (2) Given the history of the first $pr$ trials, can we predict the subsequent trials' hotel choice rate? And (3) Which modeling strategy would be best for our tasks: a non-structured classifier, a sequence model, or an attention-based approach?

The second set of questions explores the importance of the linguistic signal to our setup:
(4) Which textual features would serve our prediction model the most? Should we focus on Deep Neural Networks (DNNs) based features? Or can we also gain from hand-crafted features? And (5) Which aspects of the data are crucial for our prediction? Should we only consider the impact of the textual messages or also consider the decision-maker's behavior throughout the game?

To answer our questions, we explore different modeling strategies along three lines: (a) sequential vs. non-sequential models (which also touches on DNN based models vs. more traditional linear models); (b) text-based features learned by DNNs vs. more interpretable hand-crafted features, and (c) text-based features only, as well as their combination with behavior-based features.

We found the answers to the above questions, both encouraging and illuminating. It turns out that such action prediction in language-based persuasion games can be done effectively. Moreover, the best way to do so, is by using a mixture of a feature-based approach and a sequence neural approach. Namely, rather than learning features from the text in a plain neural approach or applying a more classical feature-based approach, we show that DNNs using relevant features allow us to obtain high-quality predictions, outperforming the baselines and the other approaches. We hence also perform interpretability analysis for our models, aiming to understand what aspects of the linguistic signals are most influential in our setup. Our use of highly effective hand-crafted features allow us to identify writing style decisions and semantic concepts that are particularly effective in driving human decisions. Another intriguing observation is that sequence models outperform the non-sequence models.

The rest of the paper is organized as follows. Section \ref{sec:previous} discusses previous work. While we are the first to address the task at hand to the best of our knowledge, we survey previous work on action prediction in machine learning, natural language processing for text-based prediction, and argumentation in multi-agent systems. Section \ref{task} defines our task, including the game definition, our prediction tasks, and how we represent our data. Section \ref{data} describes our data, including the data collection procedure, and provides a qualitative analysis. Section \ref{sec:verbal} dives into the impact of the verbal communication on the behavior of the participants in our setup. Particularly, we describe an equilibrium analysis of our game and demonstrate, through a series of additional experiments, that verbal communication increases the descriptive value of that equilibrium. Section \ref{our_approach} describes our modeling approach, including the algorithmic details of our sequential and non-sequential models, as well as the behavior and textual features. Section \ref{experiment} provides the details of our prediction experiments, including the baseline models to which we compare our approach and the evaluation measures. Sections \ref{results}, \ref{sec:interpret} and \ref{ablation_analysis} discuss our results and analyse the effect of the linguistic signal as well as of various properties of the prediction task, Finally, Section \ref{discussion} provides general discussion and conclusions.


\paragraph{Summary of Approach and Methodology}

This article introduces the study of natural language based persuasion games, namely incorporates into a persuasion game setting the use of natural language messages rather than abstract/stylized/numeric messages. In order to incorporate the effects of natural language in persuasion we need to replace abstract/stylized/numeric messages by natural language messaging. Furthermore, to more fully understand the ramifications of such change we may wish to consider the use of such messaging in multi-stage interactions (e.g. recommending many independent hotels), which better express persuasion in a realistic natural language based scenario. 
As a result a game-theoretic analysis is challenging: First, our setting is a multi-stage Pre-Bayesian game for which no results are available about equilibrium existence. In addition, the games exploit natural language messaging and the mapping from these to numeric signals is not obvious. 

To deal with these challenges we look at a simplified game, in which the various stages are played simultaneously and the players strategies are history independent. Namely, an expert (aiming to persuade) decides on his strategy for all situations, where this strategy maps the situation (hotel considered in a stage) to a numeric signal associated with the natural language review. The decision maker decides whether to accept (choose the hotel) or reject (skip the hotel) based on that numeric signal. 
Notice that by doing that we relax both the above issues, the complexity of the game and the use of natural language based signals.

For the resulting game-theoretic model we apply equilibrium analysis and derive a corresponding equilibrium. 
We then conduct behavioral experiments of the (non-simplified) persuasion setting with the natural language based setting, and also for the corresponding (non-simplified) numeric version. Surprisingly, our findings are that the behavior in the natural language based setting is close to the behavior in equilibrium more than the behavior in the numeric setting is close to the behavior in that equilibrium. Naturally, the equilibrium analysis is carried out with a stylized numeric setting, and our aim is in predicting human behavior in the original (and non-simplified) natural language based setting, and therefore, despite the above observation, we can not hope for adequate predictions in the latter. 

To overcome this, we introduce a Machine Learning (ML) approach, in which natural language messages are treated explicitly. This allows us to obtain such predictions. Together, we obtain a complete picture: Rigorous analysis that sheds light on the fact that, at least to an extent,  persuasion settings with natural language messages are behaviorally closer to “rationality” (captured by equilibrium analysis in a stylized numeric model) than numeric messages persuasion settings; but as natural language messages can not be put in a standard game-theoretic setting (and therefore does not enable rigorous analysis), we introduce an ML approach that exhibits statistically significant predictions for the (non-simplified) natural language messages persuasion setting.            


\section{Related Work}
\label{sec:previous}
We recognize three lines of work that are related to our research. This section discusses the previous works in these lines, highlights the differences between them, and summarizes our novel contributions in light of those works.

\subsection{Action Prediction in Machine Learning}
Previous work successfully employed ML techniques in service of action prediction in an ensemble of games. The first work we are aware of to use ML techniques for action prediction in one-shot strategic-form games is \citet{altman_one_shot_2006}. This work focuses on the learning of the choices made by individuals in a given game, based on population behavior in the game ensemble and the choices of the particular individual of interest in the other games. Interestingly, this approach defeats in that context leading experimental economics procedures based on cognitive models (\cite{camerer2004cognitive}; \cite{costa2001cognition}).
\citet{deep_predicting_2016} has demonstrated that DNN models trained on an ensemble of games can outperform models based on cognitive hierarchies. \citet{psyforest_plonsky_2017} have shown how psychological features can be integrated with ML techniques in order to predict human choices. The address games against nature (i.e., the choice among gambles), which are common in the psychology literature.

Overall, we identify two groups of previous work, differing in the settings they address. The first group consists of works that try to predict individuals' behavior (see, e.g., \cite{altman_one_shot_2006}, and the references therein). They represent individuals by their play in several labeled games, where all the individuals have previously played the same games. They then predict the behavior of that individual in a new, unseen game. The works in the other group are not concerned with predictions about the behavior of specific individuals, but, instead, every data point is a choice problem, e.g., a selection between two lotteries encoded by probabilities and rewards, and its label is the population statistics (e.g., \cite{hartford2016deep}; \cite{psyforest_plonsky_2017}). 

In the setting we address in this paper, we aim to predict the outcome in a new game, given information about the behavior of other players in similar (although not necessarily identical) games. Since this game is a multi-stage game, we aim to predict both the average suffix rewards and the reward in each sub-game in the suffix of the game, given the observed behavioral prefix of that game.
Beyond this difference in the game-theoretic setup, our emphasis is on introducing organic linguistic texts' strategic use in the stylized game-theoretic interaction.

\subsection{Natural Language Processing for Text-Based Prediction}

Text-based prediction tasks are ubiquitous in the Natural Language Processing (NLP) and ML literature. The most basic tasks address the predictions of properties of the text itself, including its author, topic, and sentiment \citep{joachims1999transductive,pang2002thumbs,steyvers2007probabilistic,pang2008opinion,gill2009they}. Text-based prediction have also been applied to variables that are not directly encoded in the text. One example is the prediction of what people are likely to think about a given text, e.g., predicting the citation count of a scientific paper \citep{yogatama2011predicting,sim2015utility}. Another, more ambitious, attempt is drawing predictions about real-world events based on texts that discuss related information \citep {smith2010text}. Examples of this line of work include future movie revenues based on its textual reviews \citep{joshi2010movie}, predicting risk from financial reports \citep{kogan2009predicting}, and predicting election outcomes from related tweets \citep{o2010tweets}.

Another strand of the literature on text-based prediction related to our efforts is predicting the future actions of the authors of given texts. For example, \citet{niculae2015linguistic} tried to predict actions in an online strategy game based on the language produced by the players as part of the inter-player communication required in the game. In \citet{ben2020predicting}, the authors predict an individual's action in a one-shot game based on the free text he/she provides while being unaware of the game to be played. In another study, \citet{nadavAmirRoi2020} tried to predict NBA players' in-game actions based on their open-ended interviews. However, one key difference between these tasks and our task is that in our study we aim to predict the future actions of the decision-maker who reads and uses the text although he did not produce it. Therefore, the only information we have about this decision-makers is her previous decisions. Another key difference that poses a greater challenge in our case is that we aim to predict a decision sequence, while these previous tasks did not have a sequential element.

Recently, several works studied the connection between natural language and persuasion \citep{persing2017can,carlile2018give}. \citet{wang-etal-2019-persuasion} collected a persuasion dialogue data set and predicted the persuasion strategies used in the corpus. \citet{chatterjee2014verbal} predicted speakers’ persuasiveness on housing videos of product reviews using verbal descriptors and para-verbal markers of hesitation. \citet{yang2019let} focused on advocacy requests and proposed a neural network that quantifies persuasiveness and identifies persuasive strategies. In another work, \citet{shaikh-etal-2020-examining} examined how strategy orderings contribute to persuasiveness in a loan requests data set. In contrast to these works, our work focuses on a repeated persuasion game setting, in which the expert strategy is long term, and her choice in a specific trial affects both the outcome in this trial and her reputation for the rest of the game. Another difference is that in this work we focus on the decision-makers choices, in contrast to these previous works which focus on persuasion strategies.

\subsection{Argumentation in Multi-agent Systems}

The multi-agent systems community has also conducted research related to our work, dealing with argumentation (\cite{walton2009argumentation}) and negotiation (\cite{kraus2001strategic}). Particularly, improving prediction of persuasive arguments (\cite{azaria2012strategic}; \cite{azaria2011giving}; \cite{tan_winning_arguments_2016}; \cite{rosenfeld2016strategical}) has yielded significant progress in argumentation research. Moreover, research into automated negotiations has trained automated agents to exploit human choice prediction (\cite{peled2012learning}; \cite{rosenfeld2018predicting}).

Our approach is complementary since its focus is on the task of persuasion through the use of organic linguistic texts; This is carried out in multi-stage persuasion games, extending the economics literature. We study a fundamental aspect of persuasion: Can we predict an expert (persuader) reward (i.e., the decision-maker's decisions) who aims to convince a less informed decision-maker to adopt risky alternatives while using linguistic texts? Our prediction is based only on the behavior in the prefix of the interaction between the expert and the decision-maker, the texts the decision-maker observes, and information about other experts and decision-makers' plays in different situations.
\label{related_work}

\section{Task}
\label{task}
In this section we present the main ideas of this work, the challenges, and our research questions. We aim to predict decisions within a setting of repeated expert-decision-maker persuasion games. This setting raises some challenges.

The expert in our setting observes in each trial seven reviews and their scores, and she knows the decision-maker's payment will be one of these scores. Hence, she has to decide what message (review) to send the decision-maker in order to maximize her own total payoff. This situation raises questions like: What would be a good strategy here? Should I communicate the expected payoff or should I present another statistic? The repeated aspect adds complexity, as the expert's choice in a specific trial affects not only the decision-maker's decision in this trial but also the expert's reputation for the rest of the game.

Using verbal communication introduces additional challenges. Verbal phrases add information across numerical estimates, but can also increase confusion because people interpret verbal terms in different ways (\cite{beyth1982probable}; \cite{budescu1988decisions}). For the same reason, verbal communication can also increase dishonesty, and hence may increase the experts' tendency to select inflated evaluations.

Many questions can be asked regarding this setting. This paper focuses on the decision-makers’ choices, since these decisions determine the outcome for both the expert and the decision-maker (although for the decision-maker, this is not the final payoff). Particularly, we focus on the following questions:
\begin{enumerate}
  \item Given the history of the first $pr$ trials, and the texts shown to the decision-maker in the subsequent trials, can we predict the decision made by the decision-maker in each ensuing trial?
  \item Given the history of the first $pr$ trials, and the texts shown to the decision-maker in the subsequent trials, can we predict the subsequent trials' hotel choice rate?
  \item Which modeling strategy would be best for our tasks: A non-structured classifier, a sequence model or an attention-based approach?
  \item Which textual features would serve our prediction model the most? Should we focus on DNN-based features? Or can we also gain from hand-crafted features?
  \item Which aspects of the data are crucial for our prediction? Should we only consider the impact of the textual messages or also consider the decision-maker's behavior throughout the game? 
\end{enumerate}

\paragraph{The Game} In order to implement our setup, we designed a repeated persuasion game between two players, an expert and a decision-maker, using the experimental paradigm presented in Figure \ref{exp_screen_shots} (Appendix \ref{appendix:screen}). The game consists of ten trials, played one after the other. In each trial, the expert tries
to sell the decision-maker a different hotel, by sending her textual information about the hotel. Based on this information, the decision-maker is asked to choose between 'hotel' (i.e., the risky action that provides a gain or a loss) and 'stay at home' (i.e., the safe action with a certain payoff of 0). Then, one of the seven scores is randomly selected and determined the decision-maker payoff. At the end of each trial, both participants receive the same feedback that contains 
the decision-maker's choice, the random score, and their payoffs.

\paragraph{Notations}
We now formally describe our choice prediction setup. 
Let $\hr$ be a set of hotels' reviews, and let $\hs \subset {\rm I\!R}$ be a set of hotels' scores (written in the well-known Booking.com website). Note that the scores in $\hs$ are between 2.5 and 10, and each review in $\hr$ was originally written with a related score from $\hs$. Let $\mA= \{hotel, stay\_home\}$ be the set of action choices made by the decision-makers in each trial of the experiment (which serves to define our labels). The decision-makers make these choices as a response to the textual information that the experts choose to reveal to them.


\paragraph{Prediction Task} We are interested in predicting the decisions made by the decision-makers. Specifically, given the information about the first $pr$ trials (hereinafter 'prefix') and partial information about the $sf$ (hereinafter 'suffix') following trials (where $sf=10-pr$ and $pr \in \{0, 1, ..., 9\}$), we are interested in predicting the decisions in the $sf$ subsequent trials.

More concretely, in order to represent a specific interaction with prefix size $pr$, we define the vector $v_{pr} = (hr_{1}, hr_{2},...,hr_{10}, a_{1},..., a_{pr}, rs_{1},..., rs_{pr})$ where $hr_{t} \in \hr$ is the textual information shown to the decision-maker in the $t$-th trial and $a_{t} \in \mA$ is the decision-maker's choice in trial $i$. $rs_{t} \in \hs$ is the score randomly chosen for the hotel out of the set of its review scores in the $t$-th trial, and determines the decision-maker's payoff in this trial in case of a hotel choice. Hereinafter, we refer to $rs$ as the “random score”. Given $v_{pr}$ we are interested in learning the following functions:
\begin{enumerate}
  \item $F_{trial}(v_{pr}) \in \mA^{sf}$: the decision at each trial in the $sf$ subsequent trials. Formally, our trial label in the $t$-th trial is: 
  $y_{TR_t} = \begin{cases}
  1 \ \mbox{if} \ a_t = hotel\\  
  0 \ otherwise 
\end{cases}$.
  \item $F_{ChoiceRate}(v_{pr}) \in {\rm I\!R}$: the hotel choice rate in the $sf$ subsequent trials. Formally, our choice rate label is: $y_{CR} = \frac{|\{a_t:a_t=hotel,\ \forall \ t = pr+1, ..., 10\}|} {sf} $.
\end{enumerate}

In this paper we aim to learn the above functions given the history of the first $pr$ trials and the texts shown to the decision-maker in the subsequent trials. In contrast to an online learning setup, in which after each prediction of a decision in the sequence the correct decision is revealed and the learner suffers a loss, here we adopt a batch learning setup. Hence, we do not assume the learner gets neither the correct decision nor the score randomly chosen for the hotel after predicting each suffix trial's decision.

\paragraph{Representation} To solve our prediction tasks, we map the vector $v_{pr}$ to the actual inputs of our models, using the behavioral feature space, denoted by $\mB$, and the textual feature space, denoted by $\mT$ and standing for one of the feature sources we consider in this paper (see Section \ref{attributes}). More concretely, we consider two different text representation functions: ${F_{DNN}:{\hr} \rightarrow \mT_{DNN}}$ and ${F_{HC}:{\hr} \rightarrow \mT_{HC}}$, such that a text $hr \in {\hr}$ is represented by ${F_{DNN}(hr) \in \mT_{DNN}}$ or by ${F_{HC}(hr) \in \mT_{HC}}$. In our setup, $F_{DNN}(hr)$ corresponds to DNN text representation models, while $F_{HC}(hr)$ corresponds to hand-crafted features. We also consider a representation function ${F_\mB:\mA \times \hs} \rightarrow \mB$, that maps the decision and the random feedback score
into our $\mB$ feature space, such that each decision and random score are represented by $F_\mB(a, rs) \in \mB$. Details about the texts and the feature spaces are provided in Section \ref{Attributes_Extraction}.

\section{Data}
\label{data}
In this section, we describe our experimental design and data collection process. First, we describe how we collected the participants' actions during our repeated persuasion games. Then, we provide an initial qualitative and quantitative analysis of the collected data. 

\subsection{Data Collection}
\label{data_collection}
Participants were recruited using Amazon Mechanical Turk \url{(https://www.mturk.com/)}, and the experiment was programmed using oTree (\cite{chen2016otree}). We paid the participants \$2.5 for their participation and a \$1 bonus based on their performance as described below. The average fee was \$3.07, and the experiment lasted about 15 minutes. For the experiment instructions see Appendix \ref{appendix:instructions}. 

Participants were randomly and anonymously paired, and each of them was randomly selected to be in one of two roles: Decision-maker and expert. The pairs and roles were placed at the beginning of the experiment and remained fixed until the experiment ended. 

Each experiment consists of ten trials, played one after the other. In each trial, the expert tries to sell the decision-maker a vacation in a different hotel following the experimental paradigm presented in Figure \ref{exp_screen_shots} (Appendix \ref{appendix:instructions}). The expert is not informed in advance about the hotel set or any of its properties. At each trial, she observes seven reviews written by previous visitors of one hotel, along with their scores: The reviews and their scores are presented in an ascending order of score. The expert's task is to select one of the reviews, and this review is revealed to the decision-maker as an estimation of the hotel's quality. Based only on the information sent by the expert, the decision-maker is asked to choose between 'hotel' (which provides a gain or a loss) and 'stay at home' (0 payoffs with certainty). Concretely, the decision-maker is not informed about either the score associated with the observed review or the score distribution of the hotel. The participants had limited time to make their choices. Specifically, the experts had two minutes, and the decision-makers had one minute to make their choice in each trial.

After both participants made their choices, one of the seven scores was randomly sampled from a uniform distribution over the scores. The payoff for the decision-maker, in points, from taking the hotel was this random score minus a constant cost of 8 points. Formally, the decision-maker's payoff in the $t$-th trial is:
\[DM_{payoff}(a_t) = \begin{cases}
  tr_t - 8\ \mbox{if} \ a_t = hotel\\  
  0 \ otherwise
\end{cases}\]
This cost reflects a zero expected payoff for a decision-maker who would choose the hotel option in all the ten trials. The payoff for the expert was one point if the decision-maker chose
the hotel and 0 otherwise. Formally, the expert's payoff in the $t$-th trial is:
\[Ex_{payoff}(a_t) = \ind_{a_t = hotel}\]

At the end of each trial, both participants received the same feedback that contained the decision maker's choice, the random feedback score, and their payoffs.

As an attention check, participants had to write a specific word to answer the question: “Do you have any comments on this HIT?” before clicking “I agree” at the end of the instructions. Participants who failed this attention check were excluded from the experiment, and their partners were paid for their time. The pairs of participants that passed the attention check were requested to provide personal information, such as age and gender.

At the end of the experiment, the decision-makers were
asked to take another attention check. In essence, four reviews were
presented to each decision-maker. While two of them were selected by the expert and presented to her during the experiment, the other two were not presented during the entire experiment. The decision-maker was
asked to mark the two reviews she had seen before. Decision-makers who had more than one mistake, failed the attention check and were excluded from our analysis.

The probability of obtaining the \$1 performance-based bonus was determined by the number of points the participant (expert or decision-maker) accumulated during the experiment. Specifically, we calculated the relative proportion of points that each participant earned from the maximum points she could accumulate during the experiment. If the proportion was higher than a number uniformly sampled from the [0,1] range, the participant received the bonus. We used this bonus mechanism to motivate participants to maximize their payment and, at the same time, maintain a random aspect of their final payment.

We created two separate data sets: Train-validation and test data sets. Both data sets were created using the same process, but in each of them, we used a different set of hotels (see Section \ref{Quantitative_Data_Analysis} for hotel description). In the train-validation data, we had 3548 participants, but 34\% did not pass the first attention test, and 8\% of the remaining participants left the experiment before taking the test. Next, 26\% of the remaining participants passed the attention test, but their partner did not pass it or did not take it, and hence they could not continue the experiment. This left us with 1116 participants. We created 558 pairs, but in four cases at least one participant did not meet the deadline in more than 90\% of the trials, and in 63 cases at least one participant decided to leave the experiment, and hence we filtered out these pairs. Finally, 16\% of the decision-makers that finished the experiment did not pass the second attention test or decided not to take it. These pairs were also filtered out. We thus ended up with 408 pairs (4080 trials) in the train-validation data set.

In the test data, we had 1504 participants, but 40\% did not pass the first attention test. A total of 13\% of the remaining participants left the experiment before taking the test. Next, 29\% of the remaining participants passed the attention test, but their partner did not pass it or did not take it, and hence they could not continue the experiment, which left us with 258 participants. We created 129 pairs, but in seven cases at least one participant decided to leave the experiment, and hence we filtered out these pairs. Finally, 14\% of the decision-makers that finished the experiment did not pass the second attention test or decided not to take it. These pairs were also filtered out. We thus ended up with 101 pairs (1010 trials) in the test data set.

\subsection{Quantitative Data Analysis}
\label{Quantitative_Data_Analysis}
In this section we provide an initial quantitative analysis of the data we have collected. We present the participants' properties, some of the reviews and their scores, and statistics of the decision-makers' decisions. Among the 816 participants that were included in the train-validation data, 408 were female and 408 male. The average, median, and standard deviation of participants' age were 35.5, 32, and 11.34 years, respectively. Among the 202 participants that were included in the test data, 96 were female and 106 male. The average, median, and standard deviation of participants' age were 32.8, 31, and 10.8 years, respectively.

\paragraph{The hotels and their Reviews} We use a data set with more than 500,000 reviews written in the well-known Booking.com website. To create the hotels' list for the train-validation set, we randomly select ten hotels with at least seven reviews and each review with at least 100 characters. We then randomly choose seven reviews for each hotel. To create the hotels' list for the test set, we first randomly select ten hotels (excluding the hotels in our train-validation data). We then randomly choose seven reviews for each hotel, such that each hotel is matched to one of the train-validation hotels such that both hotels have close enough score distributions (mean and median differences of up to 0.4, and maximum score difference of up to 0.8). The matching between the hotels in the train-validation and test sets is one-to-one, and this way the difference between the average over the average hotel review scores in both data sets was lower than 0.05.


We present four representative reviews from the train-validation data set in Appendix \ref{appendix:data}. 
All the experts in each data set in our setting observed the same hotels and the same reviews for each hotel. However, the test set hotels and reviews are different than those of the train-validation set (see above). At the beginning of the experiment, we randomly choose for each expert the hotel presentation order during the experiment, and the order of presentation for the positive and negative parts of each review. Table \ref{hotels_distributions} (Appendix \ref{appendix:data}) presents the hotels' score distributions in our train-validation and test sets. It shows that the mean score was 8.01 in the train-validation data and 8.06 in the test data, and six of the ten hotels in both sets had a mean score larger than 8. As mentioned in Section \ref{data}, the decision maker's payoff in points for taking the hotel at each trial is the random score minus a constant cost of 8 points. These properties of the stimuli imply that the decision maker's expected payoff from always choosing the hotel option was close to zero, and if the decision-makers choose optimally (take all the hotels with a mean score above 8), the expert's average payoff is 0.6.

The reviews differ from each other in many ways. Example properties include the lengths of their negative and positive parts, the topics mentioned in each part, the review's structure, etc. To illustrate this diversity and provide a better exposition of the textual features described in Section \ref{models}, we provide four representative reviews in Table \ref{review_examples} (Appendix \ref{appendix:data}). This table also provides the score associated with each review and the score distribution of the corresponding hotel.

\begin{figure}[htbp]
 \centering
  \includegraphics[width=1\textwidth]{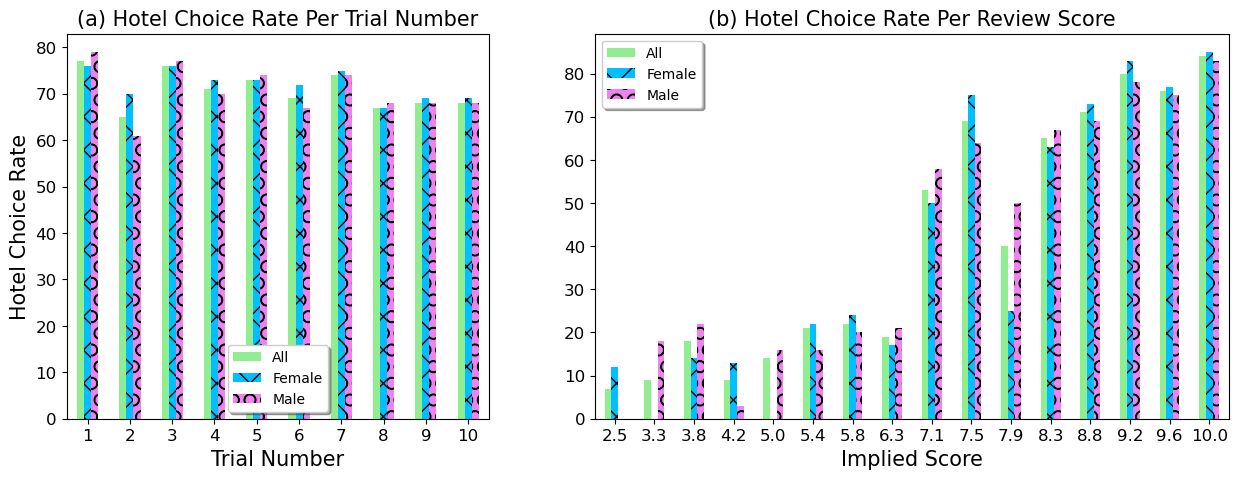}
 \caption{Hotel choice rate histograms. Figure (a) shows the percentage of decision-makers who chose the hotel option as a function of the trial number. Figure (b) shows the percentage of decision-makers who chose the hotel option as a function of the “Implied scores”: The scores associated with the reviews that were presented to the decision-makers during the experiment. }
 \label{statistics_per_round_score}
\end{figure}

\paragraph{Behavior Statistics} We now turn to discuss the statistics of the participants' behavior in the train-validation data set. This is important not only for the analysis of how participants behave in the game, but also in order to understand the data and the features we extract from the texts.
Figure \ref{statistics_per_round_score} presents the percentage of decision-makers that chose the hotel option. The left histogram presents this percentage as a function of the trial number. 
It shows that this value decreases as the number of attempts increases, but the slope is moderate. The decrease with time can imply either a better understanding of the instructions and a better inference of the hotels' quality from the content of the reviews, or a lower trust in the expert as the experiment progresses. The right histogram presents this percentage as a function of the “implied scores”: The scores associated with the reviews that were presented to the decision-makers during the experiment. This histogram demonstrates that the decision-makers tended to choose the hotel option as the review score increases. This result indicates that although the decision-makers only observed the reviews and not the scores, they could infer the quality of the hotels from the content of the reviews. Both histograms also demonstrate that there is not much of a difference between male and female participants, and hence we did not use the gender as a feature in our models.

\begin{figure}[h!]
 \centering
  \includegraphics[width=0.6\textwidth]{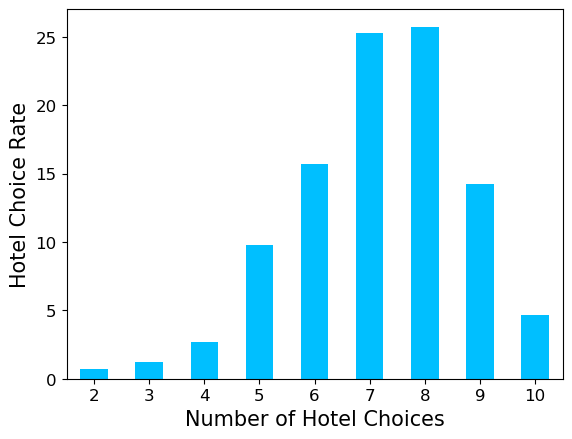}
 \caption{The tendency to choose the hotel option. For each number of trials, K, (x axis) the histogram presents the percentage of decision makers who chose the hotel K times.}
 \label{statistics_per_total_hotel_choices}
\end{figure}

As noted above, we designed our experimental paradigm such that every decision-maker that would choose the hotel option in all the ten trials would have a zero expected payoff. An optimal decision-maker would only choose the hotel option, when the mean score is above 8, i.e., in 60\% of the trials. For each number of trials, K, Figure \ref{statistics_per_total_hotel_choices} presents the percentage of decision-makers who chose the hotel K times. The histogram demonstrates that 95.3\% of the decision-makers chose the hotel option in at least half of the trials. Particularly, the average, median, and standard deviation of the total number of hotel choices were 7.18, 7, and 1.52, respectively. These results show that the decision-makers tend to choose the hotel option as expected since, in seven of the ten hotels, the median and the average score were above or very close to 8. This behavior is in line with the experimental phenomenon known as 'probability matching' (\cite{vulkan2000economist}). The results further indicate that baseline strategies that assign the average or the median hotel choice rate for every participant are very effective when it comes to predicting the overall choice rate. However, we also aim to perform this prediction correctly for populations that differ from the average participant in this aspect.

This analysis can open a direction to use Bayesian modeling that will take into consideration prior knowledge regarding the decision-makers' tendency to choose the hotel option. While we do not design such a model in this paper, we employ this knowledge in our baselines and aim to design models that outperform a baseline that predicts the average or the median hotel choice rate. 

\begin{figure}[htbp]
 \centering
   \includegraphics[width=0.6\textwidth]{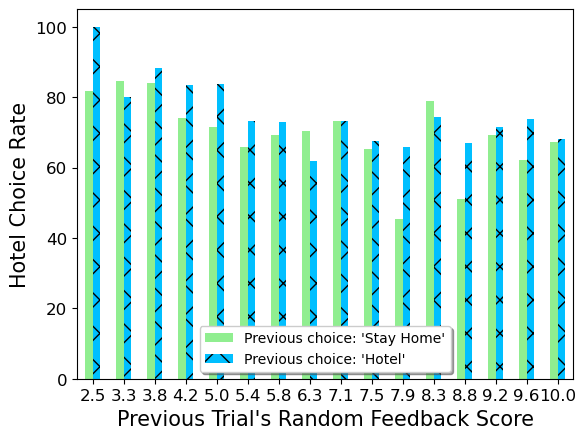}
 \caption{Decision makers' decisions as a function of previous trial decision and the random feedback score. The figure presents the percentage of decision-makers who chose the hotel option 
 both in cases where the previous decision is 'hotel' and in cases where it is 'stay at home'.}
 \label{previous_trial}
\end{figure}

Figure \ref{previous_trial} presents the decision-makers' decisions as a function of their previous choices, as well as the score that was randomly selected at the end of each trial and determines the decision maker's payoff. It shows these choice rates in the cases where the previous decision was 'hotel' and in the cases where it was 'stay at home'. The figure indicates that the decision in the previous trial and the feedback the decision-makers got, directly influence her decision in the subsequent trial. Focusing on previous trial random scores that are higher than 8 (i.e., higher than the cost of choosing the hotel option), we can infer that if the decision-makers chose the 'hotel' option and earned, they are more inclined to choose the 'hotel' option in the next trial, compared to the case where they chose the 'stay at home' option and could earn. In addition, focusing on cases where the random score selected in the previous trial was lower than 8, we can infer that the decision-makers would like to compensate for their losses in cases where they chose the hotel and lost. Generally, a previous hotel decision indicates a higher likelihood of a subsequent hotel decision again, and a higher previous random score indicates a lower probability of a hotel decision in the next trial.

In summary, our analysis shows that the decision-makers could infer the hotels' quality from the reviews' content. In addition, it indicates that the decision-makers' decision and the feedback they observe after each trial influence their decision in the subsequent trial. Finally, there is no significant difference between female and male decision-makers. These results led us to use the text the decision-makers observe at each trial, their decision and the feedback they observe in the prefix trials as features in our models. The sequential effect also calls for the application of sequential models.

In Section \ref{our_approach}, we expand our analysis to include the decision-makers' behavior as a response to the features of the text they observe. This analysis will be presented following a description of our 
hand crafted textual feature space.

As for the experts behavior, we focus on the magnitude of their exaggeration, that is, the extent to which they misrepresent the hotel by selecting an unrealistically high scoring review. To define this measure, we used the “implied scores”: The score associated with the review selected by the expert. The implied scores were transformed to “exaggeration scores” using the following equation: 
\[exaggeration = \frac{IS-SCEV}{MaxScore-SCEV}\] where IS is the implied score, SCEV is the score closest to the hotel expected value, and MaxScore is the hotel maximum score.
This measure represents the fraction of the expert's exaggeration out of the exaggeration range available to him. We consider the score closest to the expected payoff as the representation of the hotel value, since the expected payoff itself, i.e., the hotel’s average score, is not always available to the experts as the score of one of the reviews.

\begin{figure}[htbp]
 \centering
   \includegraphics[width=1\textwidth]{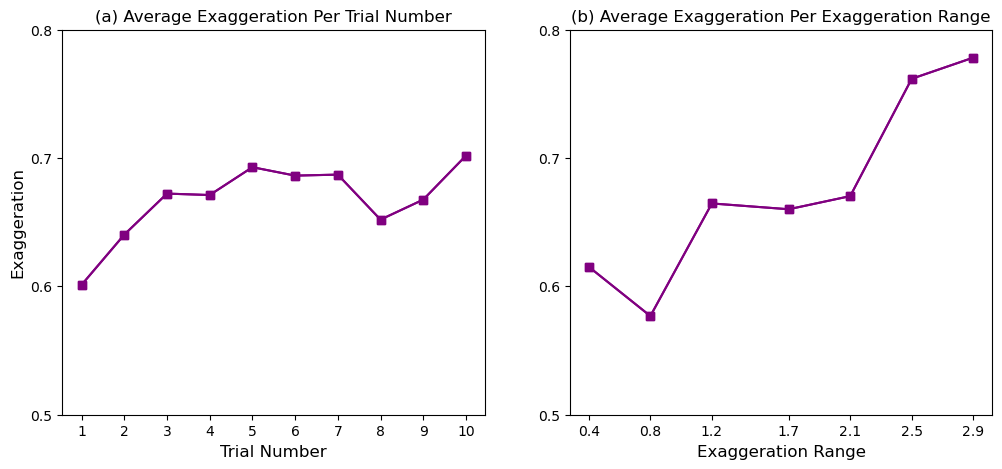}
 \caption{ Expert exaggeration Patterns. Figure (a) shows the average exaggeration as a function of the trial number. Figure (b) shows the average exaggeration as a function of the exaggeration range, i.e., the denominator of the exaggeration measure.}
 \label{figure:exaggeration_graphs}
\end{figure}

The left-hand side of Figure \ref{figure:exaggeration_graphs} presents the average exaggeration as a function of the trial number. It shows that the experts' exaggeration increases as the experiment progresses. This can be one reason for the decrease in the hotel choice rate over time, as the decision-makers may have realized that the experts increase their exaggeration as the experiment progresses. The right-hand side of Figure \ref{figure:exaggeration_graphs} presents the average exaggeration as a function of the exaggeration range, i.e., the denominator of the exaggeration measure. It shows that when the experts have more room for exaggeration, they take advantage of this opportunity and their exaggeration is more radical.

\section{Equilibrium Analysis and The Impact of Verbal Communication}
\label{sec:verbal}
The quantitative study of social and economic behavior tends to focus on simple situations in which the information available to the interacting agents includes quantitative description of the incentive structure. In their revolutionary book “the theory of games and economic behavior” \citet{morgenstern1953theory} note that this focus is a natural first step. They write (page 7): “[Economists] will have to take up first problems contained in the very simplest facts of economic life and try to establish theories which explain them and which really conform to rigorous scientific standards. We can have enough confidence that from then on the science of economics will grow further, gradually comprising matters of more vital importance than those with which one has to begin.” The current paper tries to contribute to this “growing further” process by extending the study of abstract persuasion games to address situations in which the information available to the interacting agents is incomplete, and part of the interaction uses natural language.

Bayesian persuasion assumes common-knowledge of distribution over values and one-shot messaging, and an expert that is able to obtain information about the actual value, which he can exploit in order to persuade a decision maker to take an action although its payoff is below the expected value (EV) of the distribution. In our setting there is no common-knowledge of the distribution and moreover we consider a multi-stage game in which the history observed may affect the information available and future actions, in difference to the Bayesian persuasion setting where history does not affect the decision maker choice. 

Our setup is not that of Bayesian persuasion, and in fact can be viewed as a multi-stage Pre-Bayesian game where equilibrium analysis is challenging. Aiming at some rigorous analysis we therefore consider the corresponding myopic parallel (history-independent) game. In this game the expert chooses his recommendation (signal to be transmitted) for all hotels in parallel, and similarly the decision maker decides on a strategy of when to choose (as a function of the signal received) a hotel. The decision maker's utility obtained from each chosen hotel is the hotel's expected value (recall we have seven values where one is chosen randomly under uniform distribution) minus a fixed cost (8 in our experiments). The decision maker's utility is 0 from a hotel he did not choose. The overall utility of the decision maker is the sum of his utilities from the different hotels. As for the expert, his utility is 1 (resp. 0) from a hotel the decision maker decides to choose (resp. decided not to choose), so his overall utility is just the number of hotels the decision maker decided to choose. 

Introducing some notation, let $S=\{s_1,\ldots,s_l\}$ be a set of {\em situations} (in our game $l=10$ and each situation is a hotel), where each situation $s_i$ is associated with the set of {\em signals} $\{s_i^1,\ldots,s_i^{m_i}\}$ (in our game the signals are review values and $m_i=7$ for every hotel $i$). All possible signals are taken from the interval $I=[a,b]$ (in our game it can be interpreted as $[0,10]$). 
An expert strategy space is the set of all functions of the form $f:S \rightarrow I$. 
A decision maker strategy space is the set of functions of the form $g:I \rightarrow \{choose,skip\}$. 
Given a strategy profile $(f,g)$ the expert utility in situation $s_i$, $u_e(s_i,f,g)$ is 1 if $g(f(s_i))=choose$ and 0 otherwise, 
and his total utility is $u_e(f,g)=\Sigma_{1 \leq i \leq l} u_e(s_i,f,g)$. 
The decision maker utility in situation $s_i$ is 0 if he chooses {\em skip}, and $\frac{\sum_{j=1}^{m_i}s_i^{j}}{m_i}$ - $c$ otherwise (i.e. if he chooses the hotel). Notice that  $\frac{\sum_{j=1}^{m_i}s_i^{j}}{m_i}$ is the expected value, $EV(i)$ of situation $i$, and in our setting it uniquely determines a hotel. 

The idea behind the equilibrium is to partition $S$ into sets $P_1,P2,\ldots,P_t$ where all the situations in a set $P_i$ have a common signal $s^i$, and the signals $s^i$ and $s^j$ associated with $P_i$ and $P_j$, respectively, are different. Notice that given such a partition if the expert sends the signal $s^i$ then the decision maker should accept (choose the hotel) if and only if the average of expected values in the situations associated with $P_i$ is $c$ or more, getting the expert credits as the number of elements in $P_i$ (if at least $c$) or zero (if lower than $c$). An equilibrium will be generated by finding the optimal partition (maximizing the total number of credits), while the decision maker strategy in this equilibrium is to choose the hotel if and only if the signal corresponds to a set of situations where the average of expected values is $c$ or higher. 

Notice that in a stylized setting the assumption of having signals that can effectively partition situations as above might not materialize. Nevertheless, we believe that in a world in which one can communicate with verbal statements it is much easier to find such a partition. On any case, in our particular study we were able to construct equilibria for both the train and the test sets using the above idea. In both we were able to create such a partition of the situations, where 9 of the 10 situations, which do not include the one in which signals are all lower than $c$, belong to elements of the partition with an average higher than $c$. Notice that this can be viewed as using a natural meta-strategy revealing only if the average over the expected values is above $c$ (8 in our experiments) or not, for the expert, and selecting \textit{choose} if and only if the signal is above $c$, for the decision maker. Appendix \ref{app:equ} provides
the description and the proofs of the equilibria for our training and test sets.

In the below we use the term \textit{full pooling equilibrium} when referring to the “meta-strategy” in which the expert signals \textit{above 8} for all situations that have at least one score above 8, and the decision maker strategy is choosing the hotel if an \textit{above 8} signal is received.  This leaves us with the question of when people follow this equilibrium, or at least its spirit leading to high level of persuasion. As our setting deals with language-based messaging and the scores are communicated through associated verbal descriptions, we are interested in comparing the above also to an experimental setting in which we use standard numeric messages (i.e. the associated numeric scores) as common in the game theory (both theoretical and experimental) literature. In service of that, we added a study of the same setting with numeric messages rather than natural language messages, an Only-Numerical condition. In the new study, no natural language messages are involved. Instead, the expert observed seven scores that were given to the hotel, and he is asked to select one of them to reveal to the decision-maker, which is then asked to decide whether to take the hotel or stay home.\footnote{As this study aims to answer only the question regarding the participants’ behavior, and not for predicting their choices using machine learning, we collected only 60 pairs of participants for this study.} Except for this difference, the study is the same as the study explained in Section 3.\footnote{We examined two additional conditions for this study, but in order to keep the analysis simple, we present only the results of the Only-Numerical condition. In both conditions, the expert observes the same information as in our original setup, i.e., 7 reviews alongside their scores. In the Numerical condition, the expert is asked to select one of the scores to reveal the decision-maker, while in the Verbal+Numerical condition, the expert is asked to select one review with its score to reveal to the decision-maker. Below we briefly discuss the behavior of the experts and the decision makers in these conditions.}

\begin{figure}[t!]
 \centering
   \includegraphics[width=0.5\textwidth]{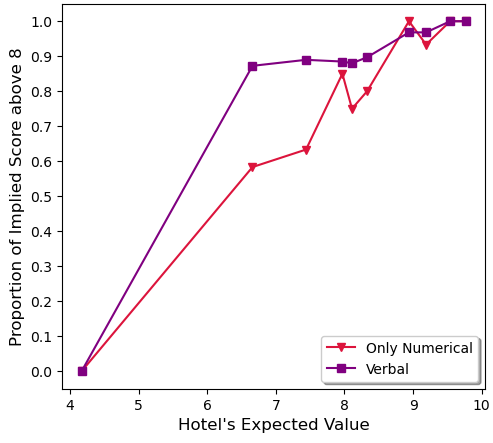}
 \caption{The proportion of implied scores above 8 as a function of the Hotel’s expected value.}
 \label{equilibrium_1}
\end{figure}

In order to analyze the participants’ behavior, we recorded the “implied scores”: The scores selected by the expert (in the Only-Numerical condition), and the score associated with the review selected by the expert (in the Verbal condition). Figure \ref{equilibrium_1} presents the proportion of implied scores above 8 at each condition as a function of the Hotel’s expected value. It shows that for hotels \#2 and \#3 (the hotels with an expected value lower than 7.97, that still have scores higher than 8), the experts’ tendency to select scores that are higher than 8 in the numerical condition is lower than in the verbal condition. Therefore, it shows the impact of the verbal review on the descriptive value of the full pooling equilibrium.\footnote{The results of the other two conditions (Numerical and Verbal+Numerical) show that around 85\% of the experts choose to reveal the decision-makers a score above 8 in hotels \#2 and \#3 (the hotels with an expected value lower than 7.97, that still have scores higher than 8). The experts in these two conditions observe both the reviews and their scores as in the Verbal condition, and since these results are almost the same as the results in the Verbal condition, they support our conclusion that the use of language increases the descriptive value of the full pooling equilibrium.}

\begin{figure}[t!]
 \centering
   \includegraphics[width=0.5\textwidth]{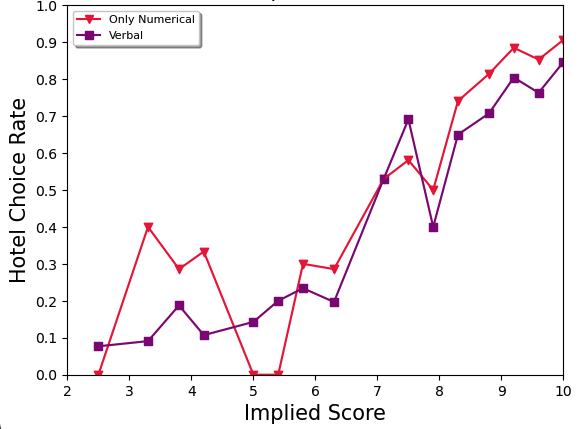}
 \caption{Hotel choice rate as a function of the implied score.}
 \label{equilibrium_2}
\end{figure}

As for the decision-makers' behavior, Figure \ref{equilibrium_2} presents the hotel choice rate at each condition as the function of the implied score, i.e., the score observed by the decision-maker (in the Only-Numerical condition) or the score of the review observed by the decision-maker (in the Verbal condition). The results show that the decision-makers in the Verbal condition behave as predicted by the full pooling equilibrium, and their tendency to select the hotel is increased as the implied score increases. One potential explanation for the high hotel choice rate as a response to implied scores between 7 and 8, is that Verbal communication is more vague than a numerical communication, and therefore, the decision-makers might interpret these reviews' scores as higher than their associated scores. In contrast, the decision-makers in the Only-Numerical condition do not follow the full pooling equilibrium, as the hotel choice rates for low implied scores are not low as expected. These rates are also higher than those of the Verbal condition. The difference between the hotel choice rates for implied scores lower than 8 is significant $(t(466)=2.68,p=0.007)$, as is the difference for implied scores higher than 8 $(t(466)=2.21,p=0.027)$.\footnote{As for the decision-makers' behavior in the remaining two conditions, the results show that presenting the decision-makers with only numerical values (in the Numerical condition) increases their tendency to select the hotel when the implied score is lower than 8, and they do not follow the full pooling equilibrium. The results of the Verbal+Numerical condition show that presenting the verbal evaluations of the hotel led the decision-makers to follow the equilibrium. Specifically, their tendency to select the hotel when the implied score is lower than 8 was lower than in the Numerical and Only-Numerical conditions. The results show that the combination of the verbal and numerical evaluations is beneficial for the decision-makers' understanding of the hotel value, as the hotel choice rates for implied scores between 7 and 8 were lower than those of the Verbal condition.}

The analysis of both players' behavior emphasizes the impact of verbal communication on the descriptive value of the full pooling equilibrium. Both the experts and the decision-makers follow the equilibrium when verbal communication is used, but they do not follow it when no natural language messages are involved. The results indicate that language has an impact on the players' behavior in our setup. This paper is hence focused on the exploration of the linguistic signal when predicting human behavior in our game.

\section{Our Approach}
\label{our_approach}
We will next describe our approach. In order to answer our research questions described in Section \ref{task}, our approach is based on two steps: In the first step, we map each vector $v_{pr} = (hr_{1}, hr_{2},...,hr_{10}, a_{1},..., a_{pr}, rs_{1},..., rs_{pr})$ to a feature vector using our textual feature sets ($\mT$) and our behavioral feature set ($\mB$). In the second step, we design non-structured classifiers, sequence models and attention-based models
that learn the two functions we discussed in Section \ref{task}: $F_{ChoiceRate}$ and $F_{trial}$. In order to predict the hotel choice rate, we consider three modeling strategies: (a) direct prediction of the choice rate; (b) prediction of the decision at each trial, i.e., designing models that learn the $F_{trial}$ function, from which the choice rate can be derived; and (c) learning of these two functions jointly. The decision sequence is considered in both the representation and the modeling steps of our approach.


\paragraph{Data Representation} In the first step, we first map each review $hr \in \hr$ to the features that represent it in our two text feature spaces (i.e., $\mT_{HC}$ and $\mT_{DNN}$) using our two functions ($F_{HC}$ and $F_{DNN}$). In addition, we map each pair $(a, rs) \in \mA \times \hs$ of a decision and a random score to our behavioral feature space, $\mB$. We then use these functions to map each vector $v_{pr} = (hr_{1}, hr_{2},..., hr_{10}, a_{1},..., a_{pr}, rs_{1},..., rs_{pr})$ into the feature vectors that will serve as our models' inputs. The features are explained in detail in Section \ref{Attributes_Extraction}, and the mapping of $v_{pr}$ into our feature space is described in Section \ref{models}.

\paragraph{Models} In the second step, we want to learn the two functions we defined in Section \ref{task}: $F_{ChoiceRate}$ and $F_{trial}$. We explore different modeling strategies, particularly sequential models that learn to predict the decision in each trial of the suffix, sequential and non-sequential models that learn to directly predict the hotel choice rate in the suffix, and joint models that jointly learn to predict both the choice rate and the decision in each trial. We now elaborate on these two steps.

\subsection{Feature Sets}
\label{Attributes_Extraction}
In this section, we describe our three feature sets: the $\mB$ feature set that represents the decision-maker's behavior in the prefix, as well as $\mT_{HC}$ and $\mT_{DNN}$ that represent the texts. The final representation of each vector $v_{pr}$ is the concatenation of the textual features and the behavioral features and is different for each model. The specific input for each model is described in Section \ref{models}.

\subsubsection{Behavioral Features ($\mB$)} Our task deals with a communication setup. We are trying to understand if we can predict future decision-maker behavior from the information that (a) was available to her and (b) can be observed by an external spectator. Specifically, what was presented to the decision-maker, what decisions did she make previously and what feedback did she observe after her decisions. Therefore, we do not encode information that was not available to the decision-maker, such as the score distribution from which her payoff is drawn or the score of the review that was selected by the expert. 

Here, we describe our behavioral feature space, $\mB$. Specifically, we map each pair $(a, rs) \in \mA \times \hs$ of a decision and a random score that determines the decision-maker's payoff to the following eight features:

\begin{enumerate}
  \item Decision: a binary feature with the decision-maker's choice at the current trial, i.e., $a \in \mA$.
  \item Random score: three binary features that indicate whether the random score ($rs$) is lower than 3, between 3 to 5, or higher than 8. Note that this random score determines the decision-maker's payoff, in case of a hotel choice.
  \item Chose and lose (cl): a binary feature that indicates whether the decision-maker chose the hotel option and lost. Formally, 
  $cl = \begin{cases}
  1 \ \mbox{if} \ a = hotel \ and \ rs < 8\\  
  0 \ otherwise
  \end{cases}$.
  \item Did not choose and could lose (nccl): a binary feature that indicates whether the decision-maker did not choose the hotel option and could have lost, had she chosen it. Formally, 
  $nccl = \begin{cases}
  1 \ \mbox{if} \ a = stay\_home \ and \ rs < 8\\  
    0 \ otherwise
  \end{cases}$.
  \item Chose and earned (ce): a binary feature that indicates whether the decision-maker chose the hotel option and earned points. Formally, 
  $ce = \begin{cases}
  1 \ \mbox{if} \ a = hotel \ and \ rs \geq 8\\  
  0 \ otherwise
  \end{cases}$.
  \item Did not choose and could earn (ncce): a binary feature that indicates whether the decision-maker did not choose the hotel option and could have earned  had she chosen it. Formally, 
  $ncce = \begin{cases}
  1 \ \mbox{if} \ a = stay\_home \ and \ rs \geq 8\\  
  0 \ otherwise
  \end{cases}$.
\end{enumerate}

Since these features provide information regarding the decision and the feedback, and since we perform batch rather than online learning, we use them to describe only the prefix trials. In Section \ref{models}, we describe the way we encode these features into each of our models. Below we refer to the value of the $j$'th behavioral feature as $\mB_{j}$.

\subsubsection{Textual Features} We have so far dealt with the representation of the behavioral information, and we will now move to describe the features that represent the texts observed by the decision makers. Previous works have already modeled sequential decision making (e.g., \cite{kolumbus_noti_2019}) but have not modeled text as the basis of these decisions, and hence this is a contribution of this paper.

We focus on two sets of textual features: $\mT_{DNN}$: Features inferred by pre-trained deep contextualized embedding models, and $\mT_{HC}$: Hand-crafted features. Research into textual representations has recently made significant progress with the introduction of pre-trained deep contextualized embedding models \citep{peters2018deep,radford2018improving}. In particular, we chose the BERT pre-trained model \citep{devlin2018bert} as our text encoder since it is dominant in NLP research. We would also like to explore the value of hand-crafted, high-level semantic features for our task, because models like BERT, that are based on language modeling related objectives, may not capture the high-level semantic information encoded in these features. We first describe our two approaches for textual representation, and in Section \ref{models} we discuss how these features are integrated into our models. Below we refer to the value of $j$'th textual feature as $\mathcal{T}_{j}$.


\paragraph{BERT-based representation ($\mT_{DNN}$)} In this approach, we utilize the pre-trained BERT model as a source of text representation. BERT is a contextualized language representation model that is based on a multi-layer bidirectional Transformer architecture and a masked language model objective. We used the cased pre-trained BERT-Base model (L = 12 layers, H = 768 hidden vector size, A = 12 attention heads, P = 110M parameters), trained on the BookCorpus (800M words) (Zhu et al. 2015) and Wikipedia (2,500M words), publicly available via source code provided by the Google Research’s GitHub repository.\footnote{\url{https://github.com/google-research/bert}} We utilized the model's source code from the “HuggingFace's PyTorch Pretrained BERT” GitHub repository.\footnote{\url{https://github.com/huggingface/transformers}} BERT can handle a sequence of up to 512 tokens, and since we use relatively short texts (our longest review contains only 144 tokens), BERT fits our needs. For each review we produce an embedding vector, by extracting the vector associated with the special [CLS] token from the last hidden layer of BERT.

\paragraph{Hand Crafted Features ($\mT_{HC}$)} We define 42 binary hand-crafted textual features, aiming to capture high-level semantic information that may not be captured by DNNs like BERT.  The hand-crafted features were constructed thorough manual inspection of the reviews in the
train-validation set. We focus on the topics and sentiment words of each part of the reviews, the review length, and whether each of the parts contains an explicit sentence that encourages or discourages visitors to stay at the hotel. The features are described in Table \ref{tab:Domain_based_attributes_description}, while Table \ref{tab:Domain_based_attributes} presents the feature representation of the hotel reviews from Appendix \ref{appendix:data} (both tables are in Appendix \ref{appendix:features}).  

Some of the features make use of sentiment words. In order to adjust such sentiment words to our goal, we decided to extract these words from the reviews, instead of using a publicly available sentiment words list. In this service, three graduate students read the train-validation hotels' reviews, extracted positive and negative sentiment words, and divided them into three groups, according to their positive and negative intensity. These lists were then merged into a unified list. The groups of positive and negative sentiment words are listed in Table \ref{tab:semantic_group} (Appendix \ref{appendix:features}). 

A similar annotation process was applied to the topic features (Table \ref{tab:Domain_based_attributes_description}) in the train-validation set. These features are the only ones that could not be easily identified in the test set in a straight-forward manner, given the annotation of the train-validation set. We hence trained a simple classifier on the train-validation reviews, to predict the existence of the topic features in a given review part, and then apply this classifier to the test reviews. The classifier consists of a standard pre-trained BERT model with a classification head for each topic feature, and its training objective is a sum of cross-entropy terms, one for each topic. The accuracy of this classifier on the test set is 96\%.

\begin{figure}[h!]
  \centering
  \includegraphics[width=1\textwidth]{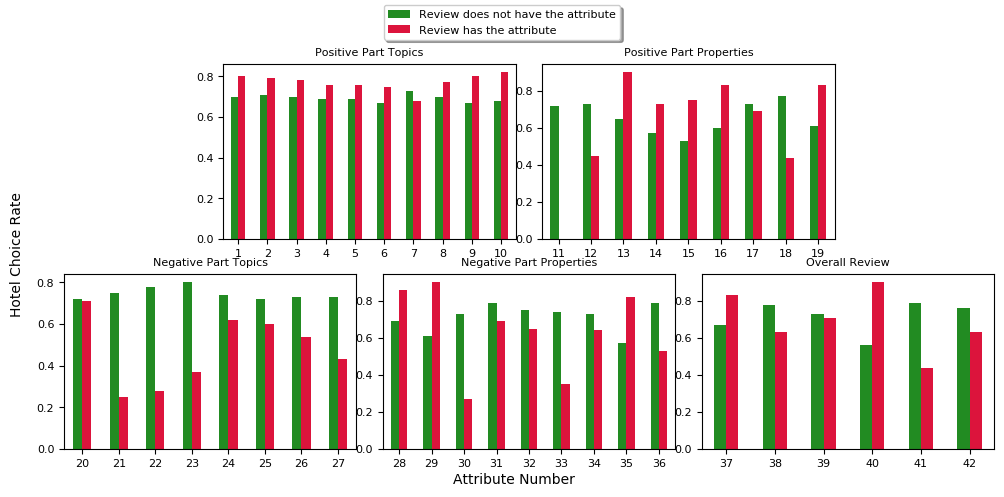}
  \caption{Hotel choice rates as a function of the hand-crafted feature values. The analysis is based on the train-validation dataset, and each histogram presents the analysis of a specific feature subset.}
  \label{enterance_rate_per_manual_features}
\end{figure}

Figure \ref{enterance_rate_per_manual_features} analyses the quality of our hand-crafted features. It shows the fraction of decision-makers in the train-validation data set that select the hotel option, in cases where the reviews they saw at each trial include or do not include each feature. For example, the numbers for feature \#11 indicate that adding a positive part to a review increases the choice rate of the hotel option by 72\%. As another example, adding a positive bottom line (as in text \#2, see Table \ref{tab:Domain_based_attributes} in Appendix \ref{appendix:features}), increases the hotel choice rate by 25\% percent, as indicated by the numbers for feature \#13. Comparing the choice rates of features \#14, \#15 and \#16 reveals that using words that more strongly emphasize positive aspects of the hotel increases the acceptance rate. A similar comparison of features \#31 -- \#33 reveals their negative impact on the hotel choice probability.

\label{attributes}

\subsection{Models}
\label{models}
Now that we have discussed the properties of our task and data, we are ready to present the way we predict the outcome of the human behavior in our task, i.e. learning the functions $F_{ChoiceRate}(v_{pr})$ and $F_{trial}(v_{pr})$ from Section \ref{task}. We propose models that learn each of these functions separately (denoted with -CR, for Choice Rate, and -TR, for TRial, respectively), as well as models that learn them jointly (denoted with -TRCR). We are particularly focused on two modeling aspects: Comparing sequential models to non-sequential ones, and comparing DNN-based models to models that do not apply this approach. We next provide a high level description of our models, and then proceed with more specific details.

\paragraph{$F_{ChoiceRate}()$ Models}. We implement three models for $F_{ChoiceRate}(v_{pr})$.
Two of our models are DNN-based: One employs a Long Short-Term Memory (LSTM) recurrent architecture (\cite{lstm}) for sequence processing, and the other employs the Transformer architecture (\cite{attention_2017}) for sequence-to-sequence learning with self-attention.
The third model is a Support Vector Machine (SVM) (\cite{cortes1995svm}), which will let us evaluate the power of a non-DNN and non-sequential modeling approach.

\paragraph{$F_{trial}()$ Models}. We implement two models for $F_{trial}(v_{pr})$. Note that these models address a more general task than the $F_{ChoiceRate}()$ models, since the hotel choice rate can be derived from their per-trial predictions. As for $F_{ChoiceRate}()$, one of our $F_{trial}()$ models is based on an LSTM architecture and one on the Transformer architecture. 

\paragraph{Joint modeling of $F_{ChoiceRate}()$ and $F_{trial}()$}. Multi-task Learning is an approach in which multiple learning tasks are solved jointly, by sharing related information (\cite{reichart2008multi,ruder2017overview}). As the choice rate and per-trial prediction tasks are tightly connected, we hope that solving them jointly would produce better results on each. Multi-task learning has been applied to a variety of NLP tasks, and DNNs are particularly suitable for its implementation (e.g. \cite{sogaard-goldberg-2016-deep}; \cite{rotman2018bridging}; \cite{malca-reichart-2018-neural}). We therefore implemented a model that jointly learns the $F_{ChoiceRate}()$ and $F_{trial}()$ functions. As for the $F_{ChoiceRate}()$ and $F_{trial}()$, one of our joint models is based on an LSTM architecture and one on the Transformer architecture. We next describe each of our models in detail.
  
\subsubsection{Support Vector Machines for $F_{ChoiceRate}()$ (SVM-CR)}
Here we describe our SVM regression model that predicts the hotel choice rate in a given trial suffix.
Since SVMs are not sequential models, this model considers the past sequence of texts and behaviors only through its features. We considered various possible representations and, based on development data experiments, we decided to represent the input state $v_{pr}$ by the weighted average of the prefix trials' textual features ($PW\mT$), the weighted average of the prefix trials' behavioral features ($PW\mB$), and the average text features of all the trials in the suffix ($SW\mT$). The weighted average of the prefix trials is defined such that earlier trials get lower weights.

Formally, the weighted average of the $j$-th prefix behavioral feature, $\mB_{j}$, and the weighted average of the $j$-th prefix textual feature, $\mathcal{T}_j$, for a prefix size $pr$, are:
\[PW\mB_j = \frac{1}{pr} \sum\limits_{t=1}^{pr} 0.8^{pr+1-t} \cdot \mB_{jt}\]
\[PW\mathcal{T}_j = \frac{1}{pr} \sum\limits_{t=1}^{pr} 0.9^{pr+1-t} \cdot \mathcal{T}_{jt}\]
where $\mB_{jt}$ is the value of the $j$-th behavioral feature in the $t$-th trial 
, $\mathcal{T}_{jt}$ is the value of the $j$-th textual feature in the $t$-th trial 
, and 0.8 and 0.9 are hyper-parameters tuned on our development data.

For example, the vector $v_{4}$ (i.e., a vector of an example with a prefix of size 4) is mapped into the concatenation of the following features:
\begin{itemize}
  \item The prefix trials' weighted $j$-th behavioral feature: $PW\mB_j = \frac{1} {4} \sum\limits_{t=1}^4 0.8^{5-t} \cdot \mB_{jt}, \forall \mB_{jt} \in \mB$
  \item The prefix trials' weighted $j$-th textual feature: $PW\mathcal{T}_j = \frac{1} {4} \sum\limits_{t=1}^4 0.9^{5-t} \cdot \mathcal{T}_{jt}, \forall \mathcal{T}_{jt} \in \mathcal{T}$
  \item The suffix trials' weighted $j$-th textual feature: $SW\mathcal{T}_j = \frac{1}{6} \sum\limits_{t=5}^{10} \mathcal{T}_{jt}, \forall \mathcal{T}_{jt} \in \mathcal{T}$
\end{itemize}



\subsubsection{Deep Neural Network Modeling}
\label{dnns}

DNNs have proven effective for many text classification tasks (\cite{kim2014convolutional}; \cite{ziser2018deep}). In this part of the paper, we provide a high level description as well as more specific details of our DNN models.

In our DNN models, each trial in the prefix is represented using its behavioral features (as described in Section \ref{Attributes_Extraction}). These features are concatenated to the trial's textual features (either its $\mT_{DNN}$ or its $\mT_{HC}$ features, or their concatenation, as described in Section \ref{Attributes_Extraction}). In contrast, since the suffix trials' behavioral features are not known, each trial in the suffix is represented only with its textual features. 

\paragraph{The LSTM Models} These models belong to the family of Recurrent Neural Networks (RNNs), which can process variable length sequences. We hypothesize that since our data involve multiple trials, and based on our analysis described in Section \ref{Quantitative_Data_Analysis} where we show a sequential effect in the decision-making process, a sequential model could capture signals that non-sequential models cannot.

LSTM is an RNN variant designed to handle long-distance dependencies, while avoiding the vanishing gradients problem. It has shown very useful in sequence modeling tasks in NLP, such as language modeling (\cite{sundermeyer2012lstm}), speech recognition (\cite{greff2016lstm}) and machine translation (\cite{wu2016google}). We describe our LSTM models below, focusing on their labels, input vectors and architectures.

\begin{figure}[t]
  \centering
  \includegraphics[width=1\textwidth]{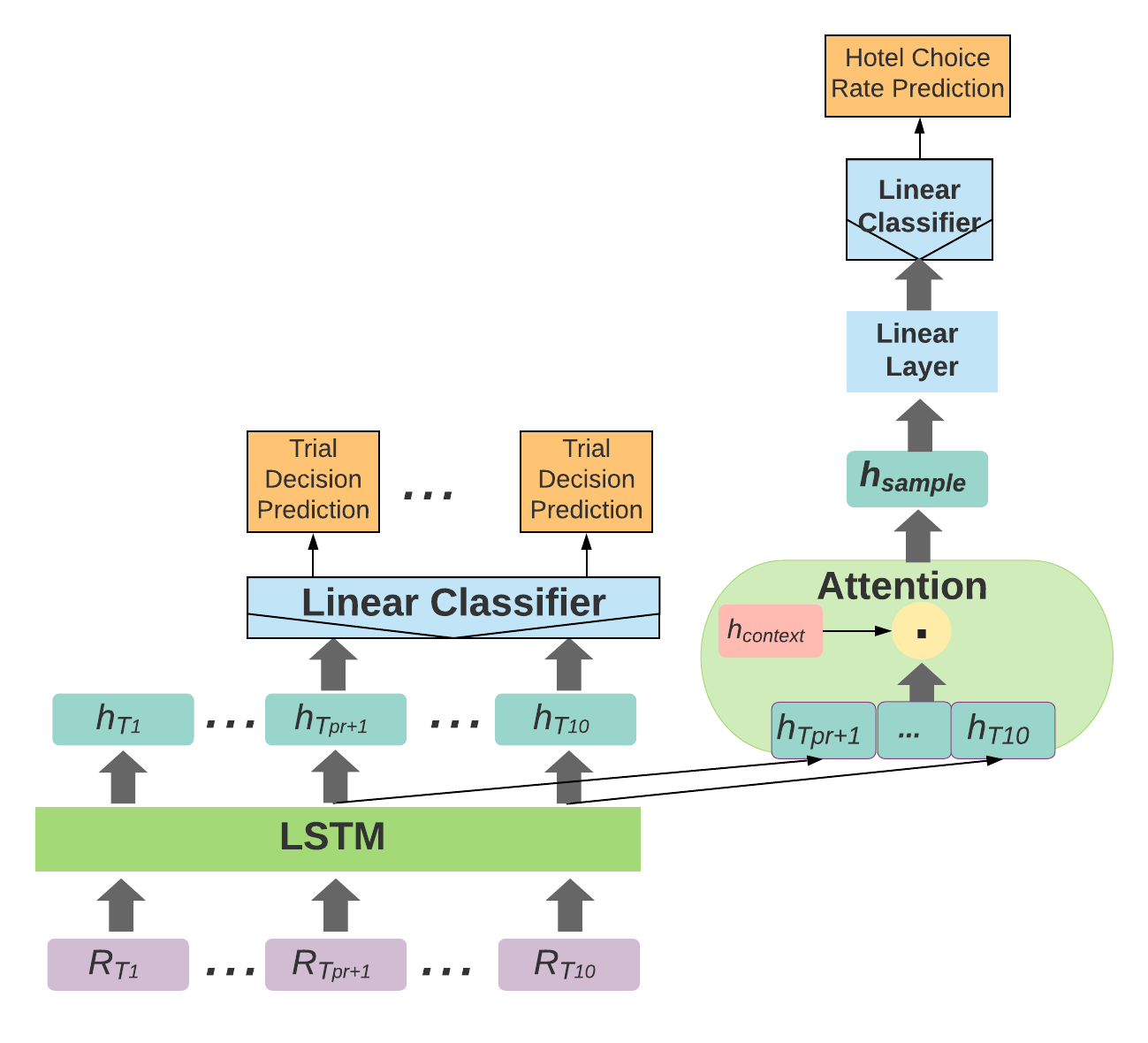}
  \caption{The LSTM-based models. $pr$ denotes the example prefix size and $R_{T_{i}}$ denotes the representation vector of the $i$-th trial. The left part is the LSTM-TR model, the right part is the LSTM-CR model, and the complete figure presents the joint LSTM-TRCR model.}
  \label{use_raisha_lstm_based_model}
\end{figure}

We have considered various LSTM-based models and multiple approaches for mapping the input $v_{pr}$ as these models' input. Each input $v_{pr}$ is represented with a sequence of feature vectors, such that each feature vector represents one trial (either a prefix or a suffix trial; the feature vectors of each prefix and suffix trial are described above). We next describe the best model version based on our development data results, an illustration of the architecture is provided in Figure \ref{use_raisha_lstm_based_model}.

\textbf{\textit{LSTM-CR}}. This is the LSTM model that predicts the hotel choice rate in the suffix. Figure \ref{use_raisha_lstm_based_model} (right) provides a description of this architecture. 
The LSTM is sequentially fed with the prefix and suffix trials' representations, one trial at a time. The suffix trials' hidden vectors are fed into a dot product attention layer, followed by two linear layers with a dropout layer and a ReLU activation function, in order to predict the hotel choice rate in the suffix trials. The model applies the mean squared error (MSE) loss as implemented in the PyTorch.nn module:\footnote{\url{https://pytorch.org/docs/stable/generated/torch.nn.MSELoss.html}}
\[MSE = \frac{1}{batch}\sum\limits_{i=1}^{batch}(\hat{y}_{CR_i} - y_{CR_i})^2\]
where $batch$ is the size of the training batch (in the stochastic optimization process), and $\hat{y}_{CR_i}$ and $y_{CR_i}$ are the predicted and the gold hotel choice rates in the $i$-th example of the batch, respectively.

\textbf{\textit{LSTM-TR}}. This is the LSTM model that predicts the decision in each suffix trial. The LSTM-TR architecture is described in the left side of Figure \ref{use_raisha_lstm_based_model}. The result of this model can also be averaged in order to get the hotel choice rate in the suffix trials.

The LSTM is sequentially fed with the prefix and suffix trials' representations, one trial at a time.
Each hidden state of the suffix trials is fed into a dropout layer followed by a linear layer with a ReLU activation function, in order to predict the label for each suffix trial. The loss function of this model is the sequence cross-entropy (SCE), as implemented in the AllenNLP software package:\footnote{\url{https://github.com/allenai/allennlp/blob/master/allennlp/nn/util.py}}
\[SCE = \frac{1}{batch}\sum\limits_{i=1}^{batch}\frac{\sum\limits_{j=pr+1}^{10}-(y_{TR_{ti}}\cdot log(p_{ti}) + (1-y_{TR_{ti}})\cdot log(1-p_{ijs}))}{sf}\]
where $batch$ is the size of the training batch (in the stochastic optimization process), $pr$ is the prefix size, $sf$ is the suffix size, $p_{ti}$ is the predicted probability that the $t$-th trial of the $i$-th example of the batch is hotel, $a_{ti}$ is the decision in the $t$-th trial of the $i$-th example of the batch and $y_{TR_{ti}} \in \{0, 1\}$ is the choice of the $t$-th example of the batch in the $i$-th trial, such that 
$y_{TR_{ti}} = \begin{cases}
  1 \ \mbox{if} \ a_{ti} = hotel\\  
  0 \ otherwise 
\end{cases}$.
\\\\
\textbf{\textit{LSTM-TRCR}}. This model jointly learns to predict the decisions made by the decision maker in each trial, and the hotel choice rate. The LSTM-TRCR architecture, a combination of the above LSTM-TR and the LSTM-CR, is described in Figure \ref{use_raisha_lstm_based_model}.

Since the choice rate and the trial labels of each example are strongly related, such that the hotel choice rate label is an average of the trial labels, we augment the above losses with a loss term that is aimed to minimize the squared distance between the predicted choice rate and the average of the individual trial predictions. For this purpose, we calculated the averaged trial predictions using the argmax value of a softmax layer that is fed with the sequence of trial prediction. Formally, given the above notation, and defining $\hat{y}_{TR_{ti}} \in [0, 1]$ to be the prediction of the $t$-th trial of the $i$-th example, we define the mean squared trial-choice rate error (MSTRCRE):
\[MSTRCRE = \frac{1}{batch}\sum\limits_{i=1}^{batch} (\hat{y}_{CR_i} - \frac{1}{sf}\sum\limits_{j=pr+1}^{10} \hat{y}_{TR_{ti}})^2\]
Finally, we define the trial-choice rate loss (TRCRL) as the weighted average of three losses, MSE, SCE and MSTRCRE: 
\[TRCRL = \alpha \cdot MSE + \beta \cdot SCE + \gamma \cdot MSTRCRE\]
where $\alpha$, $\beta$ and $\gamma$ are hyper-parameters.

\paragraph{The Transformer Models}. Another neural network model that has proven to be especially effective for many natural language processing tasks is the Transformer (\cite{attention_2017}). The Transformer has shown very useful in various NLP tasks, including machine translation (\cite{attention_2017}, \cite{shaw-etal-2018-self}) and speech recognition (\cite{dong2018speech}), among many others. The Transformer is a sequence-to-sequence model that consists of an encoder and a decoder. In our case, the encoder maps the prefix trials' input sequence to a sequence of continuous representations. Given these representations and an input sequence of the suffix trials, the decoder then generates an output sequence of the suffix trials' representations fed to our model's next layers to generate the output predictions. 
Below we describe our Transformer models: Their labels, input vectors and architectures.

\begin{figure}[h]
  \centering
  \includegraphics[width=1\textwidth]{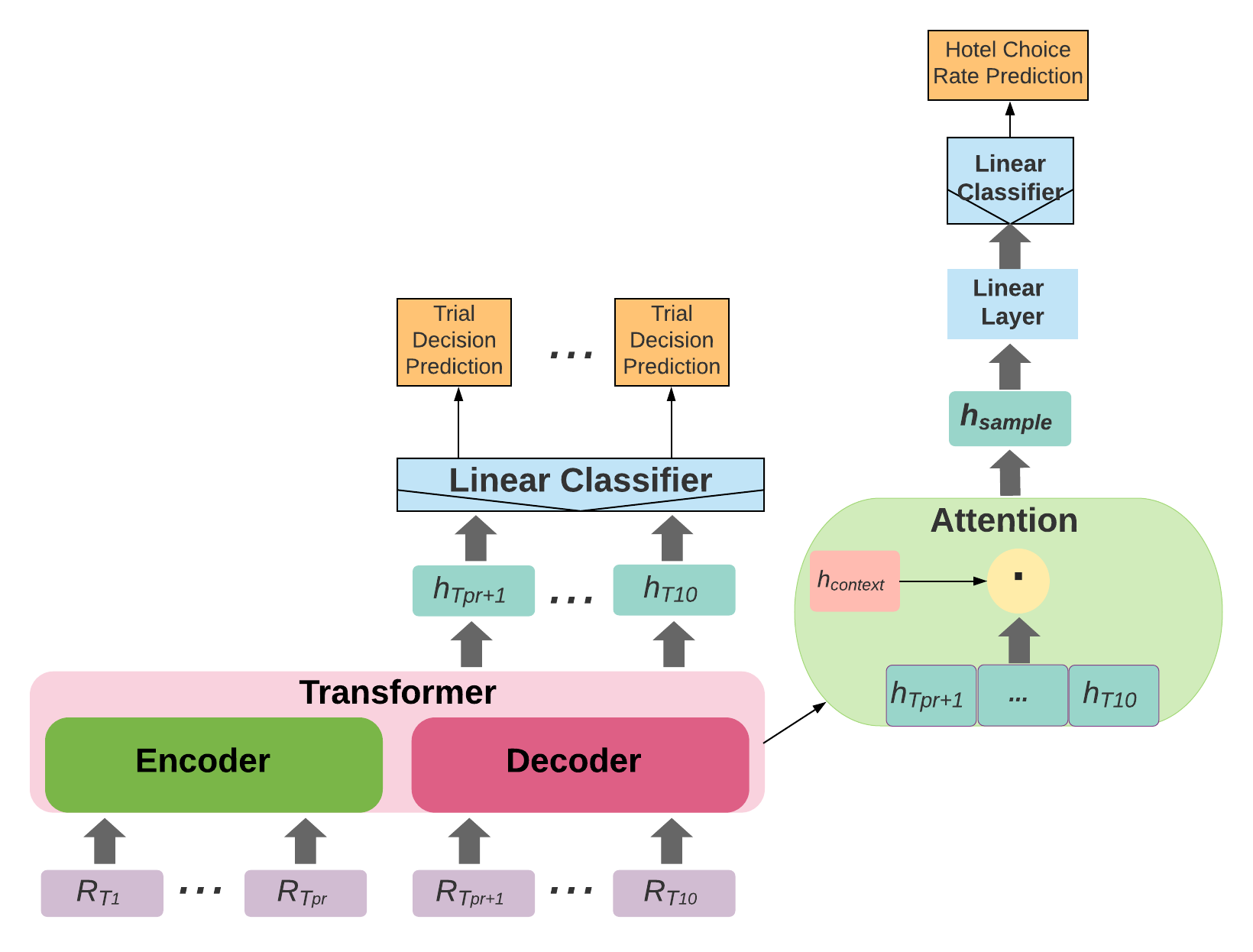}
  \caption{The Transformer-based models. $pr$ denotes the prefix size of the sample, $R_{T_{i}}$ denotes the representation vector of trial $i$, and $h_{context}$ is randomly initialized and learned jointly with the attention weights during the training process. The left part is the Transformer-TR model, the right part is the Transformer-CR model, and the entire figure stands for the joint Transformer-TRCR model.}
  \label{transformer_based_model}
\end{figure}

By implementing Transformer-based models, we aim to model each input $v_{pr}$ as two sequences: A sequence of the prefix trials and a sequence of the suffix trials, so that to model our task as a translation of the prefix trials to the decisions in the suffix trials. The representations of both the prefix and suffix trials are described above. Since the model's input consists of two sequences, we did not feed it with examples with $pr=0$. See Figure \ref{transformer_based_model} for an illustration of the model architecture.

\textbf{\textit{Transformer-CR}}. This is the Transformer model that predicts the hotel choice rate in the suffix, and its architecture is described in the right side of Figure \ref{transformer_based_model}. The Transformer is fed with two sequences as described above, and its output is a sequence of $sf$ hidden vectors. These hidden vectors are fed into a dot product attention layer, followed by two linear layers with a dropout layer and a ReLU activation function, in order to predict the hotel choice rate in the suffix trials. The loss function of this model is the MSE loss described above.

\textbf{\textit{Transformer-TR}}. The Transformer model that predicts the decision in each suffix trial, and its architecture is described in the left side of Figure \ref{transformer_based_model}. The output of this model can also be averaged to get the hotel choice rate in the suffix trials. The Transformer is fed with two sequences as described above, and its output is a sequence of $sf$ hidden vectors. Each hidden vector is fed into a linear layer with a dropout layer and a ReLU activation function in order to predict the label for each suffix trial. The loss function of this model is the SCE loss described above.

\textbf{\textit{Transformer-TRCR}}. This is the Transformer model that jointly predicts the per-trial decision and the overall hotel choice rate. This model is a combination of the two models described above: The Transformer-TR and the Transformer-CR, and its architecture is described in Figure \ref{transformer_based_model}. The loss function of this model is the TRCRL loss described above.

\section{Experiments}
\label{experiment}
Recall from Section \ref{data_collection} that we have collected a train-validation set of 408 examples and a separate test set of 101 examples. Each example in each set consists of a ten-trial game, and the sets differ in their hotel (and hence also review) sets. We break each example in each of the sets into 10 different examples, such that the first $pr \in \{0,1, \ldots, 9\}$ trials serve as a prefix and the remaining $sf = 10 - pr$ trials serve as a suffix. This yields a total of 4080 training examples and 1010 test examples.

\subsection{Models and Baselines}
\label{models_baselines}
We consider the following models (described in further detail in Section \ref{models}): SVM-CR, 
LSTM-CR, LSTM-TR, LSTM-TRCR, Transformer-CR, Transformer-TR and Transformer-TRCR, where −CR, −TR and −TRCR denote model variants for hotel choice rate, per-trial decisions and joint TR and CR predictions, respectively.

\paragraph{Research Questions}
Recall our five research questions from Section \ref{task}. Our experiments are designed to compare different modeling strategies in learning one of two functions: (1) $F_{trial}()$ and (2) $F_{ChoiceRate}()$. 
In addition, they are designed to (3) compare different modeling strategies: A non-structured classifier, a sequence model and an attention-based approach. Our experiments are also designed to (4) compare different types of text-based features, and (5) compare between models with text-only and both text and behavior features, in order to explore whether these features have a complementary effect.

\paragraph{Model Variants and Comparisons}
We perform two sets of comparisons between variants of our models. The comparisons are designed so that to allow us to answer questions \#1-\#3 about the prediction capacity of our models and their optimal structure, while directly addressing our questions about the optimal textual representation (question \#4) and the complementary effect of textual and behavioral features (question \#5). 

In the first set of comparisons, related to question \#4, we consider three variants of each of the models, such that all models use the entire set of features (both textual and behavioral features), but they differ from each other in their textual features: A variant that employs only the $\mT_{DNN}$ (BERT-based) features, a variant that uses the $\mT_{HC}$ (hand crafted) features, and a variant that employs both sets. In the second set of comparisons, related to question \#5, we consider two variants of our models: A variant with textual features only (for both the prefix and the suffix trials), and a variant with both the textual and the behavioral features for all trials, similarly to the first set of comparisons. Clearly, these comparisons will also provide us with answers to questions \#1-\#3.

Note that since we use batch learning models, we use the behavioral features to represent only the prefix trials as they are not known for the suffix trials. Also, since our main focus of this paper is on the use of texts in a persuasion games setup, we do not consider a variant of our model with behavioral features only.

\paragraph{Baselines}
Similar to our models, the baselines we consider predict either the hotel choice rate or the decision at each trial (and derive the hotel choice rate from the per-trial prediction).

As choice rate baselines, we consider the \textit{Average} (AVG) and the \textit{Median} (MED) baselines. These baselines assign to each example in the test set the average or the median hotel choice rate label, respectively, as computed over the training set examples:
\[AVG = \sum\limits_{i=1}^{|\mT|} \frac{y_{CR_i}}{|\mT|}\]
where $y_{CR_i}$ is the hotel choice rate label of the $i$'th example in the training set $\mT$. Let $ChoiceRates = \{y_{CR_i} | i\in \mT\}$ be the set of the hotel choice rate labels of all the training set examples, the MED baseline assigns the following label to test-set example:
\[MED = \frac{1}{2} (ChoiceRates_{\lfloor{(|\mT|-1)/2}\rfloor} + ChoiceRates_{\lceil{(|\mT|+1)/2}\rceil})\]

We also consider per-trial baselines. The first is the strong \textit{Majority Vote Classifier} (MVC) that assigns to each trial of each test set example the majority decision across all training set trials. Formally, let $y_{TR_{ti}} \in \{0,1\}$ be the trial label of the $t$-th trial, of the $i$-th training set example with prefix of size $pr_i$ and suffix of size $sf_i = 10 - pr_i$, and let 
\[avg\_TR\_label = \frac{\sum\limits_{i=1}^{|\mT|} \sum\limits_{t=pr_i+1}^{10} y_{TR_{ti}}} {\sum\limits_{i=1}^{|\mT|} sf_i}.\]
be the average of these labels. The MVC baseline then assigns to each trial in the sequence of each test set example the following prediction:
\[MVC = \begin{cases}
  1 \ \mbox{if} \ avg\_TR\_label \geq 0.5\\  
  0 \ otherwise 
\end{cases}.\]
Note that this baseline assigns the same label to all the trials of all the test set examples. While one might also like to consider a per-trial majority vote baseline, we noticed that the majority votes of all trial numbers (from 1 to 10) were identical (that is, take the hotel) which makes this baseline identical to our MVC. 


In addition, we consider an \textit{Expected Weighted Guess} (EWG) baseline. For this baseline we compute the expected quality of the predictions of a stochastic per-trial classifier, which assigns every test-set trial with one of the two possible labels according to the unigram probability of that label in the training set (when computing this probability we do not distinguish between trial numbers). Note that this is a theoretical classifier that we use in order to put our results in context. Since this classifier is not deterministic, it cannot be applied in practice. We evaluated this theoretical classifier by drawing 5000 assignments for the test-set trials and averaging the resulting values of our evaluation measures. 



\subsection{Evaluation Measures}
\label{evaluation_measures}
We consider three evaluation measures in order to analyze the multiple aspects of our task. 

\paragraph{Trial-level measures}
To evaluate the performance of the various models in predicting the decision at each trial, $F_{trial}()$, we consider the Accuracy-Per-Trial measure, defined as the fraction of test-set trials that are correctly labeled by the algorithm. More specifically, let $y_{TR_{ti}}$ be the trial label of the $t$-th trial, of the $i$-th training set example with prefix of size $pr_i$ and suffix of size $sf_i = 10 - pr_j$, and let $\hat{y}_{TR_{ti}}$ be the predicted trial label. The Accuracy-Per-Trial measure is:
\[Accuracy-Per-Trial = \frac{\sum\limits_{i=1}^{|\ts|} \sum\limits_{t=pr_i+1}^{10}\ind_{y_{TR_{ti}}=\hat{y}_{TR_{ti}}}} {\sum\limits_{i=1}^{|\ts|} sf_i}\]

We also compute the Macro Average F1-score: We compute the F1-score of each label and report the average of the resulting values. More specifically, let $y_{TR_{ti}}$ be the trial label of the $t$-th trial, of the $i$-th training set example with prefix of size $pr_i$, and let $\hat{y}_{TR_{ti}}$ be its predicted trial label. For each class $i \in ('hotel', 'stay\_home')$ we compute:
\[Precision_i = \frac{\sum\limits_{i=1}^{|\ts|} \sum\limits_{t=pr_i+1}^{10}\ind_{y_{TR_{ti}}=\hat{y}_{TR_{ti}}=i}} {\sum\limits_{i=1}^{|\ts|} \sum\limits_{t=pr+1}^{10} \ind_{\hat{y}_{TR_{ti}}=i}}, Recall_i = \frac{\sum\limits_{i=1}^{|\ts|} \sum\limits_{t=pr_i+1}^{10}\ind_{y_{TR_{ti}}=\hat{y}_{TR_{ti}}=i}} {\sum\limits_{i=1}^{|\ts|} \sum\limits_{t=pr_i+1}^{10} \ind_{y_{TR_{ti}}=i}}\]
\[F1_i = \frac{2 \cdot Recall_i \cdot Prediction_i}{Recall_i + Prediction_i}.\]
In other words, $Precision_i$ is the fraction of the examples belonging to class $i$ from those classified as class $i$, while $Recall_i$ is the fraction of the examples from class $i$ that are classified as class $i$. The $F1_i$ score is the harmonic average of the precision and recall of the $i$-th class. The Macro Average F1-score is:
\[Macro-F1 = \frac{1}{2}\sum\nolimits_{i \in ('hotel', 'stay\_home')} F1_i \] i.e., it is the average of the class-based F1-scores, with equally weighted classes.

\paragraph{Choice rate measures}
To evaluate the performance of the various models in predicting the choice rate, $F_{ChoiceRate}()$, we employ the Root Mean Square Error (RMSE) measure:
\[RMSE = \frac{1}{|\ts|}\sqrt{\sum\limits_{i=1}^{|\ts|}(\hat{y}_{CR_i} - y_{CR_i})^2}\]
where $y_{CR_i}$ is the choice rate label of the $i$-th example in the test set $\ts$, and $\hat{y}_{CR_i}$ is the predicted choice rate of that example.

Since most of the participants in our experiment behave similarly to the average participant (see details in Section \ref{Quantitative_Data_Analysis}), the RSME measure would not indicate that a model fails in capturing behaviors that deviate from the average behaviour. We hence perform a bin analysis, mapping the choice rates into four bins:
\begin{itemize}
  \item Bin 1: $choice \ rate < 0.25$
  \item Bin 2: $0.25 \leq choice \ rate < 0.5$
  \item Bin 3: $0.5 \leq choice \ rate < 0.75$
  \item Bin 4: $choice \ rate \geq 0.75$
\end{itemize}
We then compute the Macro Average F1-score over these bins. More specifically, let $y_{bin_i}$ be the bin label of the $i$-th example in the test set $\ts$, and let $\hat{y}_{bin_i}$ be the predicted bin of that example. For each bin $j \in (1, 2, 3, 4)$ we compute:
\[Precision_j = \frac{\sum\limits_{i=1}^{|\ts|} \ind_{\hat{y}_{bin_j}=y_{bin_j}=j}}{\sum\limits_{i=1}^{|\ts|} \ind_{\hat{y}_{bin_j}=j}}, Recall_j = \frac{\sum\limits_{i=1}^{|\ts|} \ind_{\hat{y}_{bin_i}=y_{bin_i}=j}}{\sum\limits_{j=1}^{|\ts|} \ind_{y_{bin_i}=j}}\]
\[F1_j = \frac{2 \cdot Recall_i \cdot Prediction_i}{Recall_i + Prediction_i}.\]
The Macro Average F1-score is then:
\[Bin-Macro-F1 = \frac{1}{4}\sum\limits_{j=1}^4 F1_i \] 

Note that we evaluate the choice rate models using the RMSE and the bin analysis measures only. In contrast, we evaluate the per-trial and the joint models using all our evaluation measures, because the choice rate can be derived from the predictions of the per-trial models.


\subsection{Cross Validation}
We employ a six-fold cross-validation protocol in order to tune the hyper-parameters of each model. For this purpose, we split the 408 (expert, decision-maker) pairs of the train-validation set into six subsets, such that each subset consists of 68 pairs. As described above, each decision sequence is translated into ten examples, each with a different prefix size, resulting in 680 examples in each subset. In each fold, we select one subset for development and the remaining five subsets serve for training. Each model is then trained on the training set, and its hyper-parameters are tuned on the development set so that to minimize its RMSE loss. The tuned models are then applied to the held-out test set. Finally, for each model we report the average results across the six folds.

\subsection{Statistical Significance Testing}

We would like to make sure that the differences between the models we evaluate are statistically significance. Following \citet{dror2018hitchhiker} we use the McNemar’s test \citep{mcnemar} to test the significance of our trial-level results and of the results of the bin analysis we perform over the choice rate results. We also use the two-tailed t-test to test the significance of the square error choice rate results. These tests are used in order to ensure the significance of the differences between chosen model pairs. Since we employ a cross-validation protocol, we compute the p-values for each of the folds, employ the Bonferroni correction to examine the significance across folds, and report the number of folds with a statistically significant effect \citep{dror2017replicability}. In what follows we consider a result to be statistically significant if $p < 0.05$. \footnote{We note that it is customary in the NLP literature to perform either cross-validation or statistical significance testing. Here we apply an extra level of significance analysis and compute the significance across folds with the conservative Bonferroni correction. For further discussion of statistical significance testing in NLP when cross-validation is performed we refer the reader to  \citet{dror2018hitchhiker} and \citet{dror2020statistical}.}


\subsection{Hyper-Parameter Tuning}
We next describe the hyper-parameters of each of the models.

\paragraph{SVM} We use the standard Support Vector Regression (SVR) model of the sklearn package to predict the hotel choice rate.\footnote{\url{https://github.com/scikit-learn/scikit-learn/tree/master/sklearn/svm}} We use the default values for all the model hyper-parameters and tune the type of kernel function (rbf, linear, polynomial) as well as the polynomial degree of the kernel function (3, 5, 8). 


\paragraph{DNNs} For all DNNs, we use ReLU as the activation function for all internal layers, and we tune the dropout parameter (0.0, 0.1, 0.2, 0.3), such that the same dropout parameter was used in the LSTM and Transformer models, as well as in the linear layers placed on top of these models. Training is carried out for 100 epochs with early stopping, and a batch size of 10 in the LSTM-based models and 9 in the Transformer-based models. Each batch consisted of all the examples of one decision-maker. We use a different batch size for each model, since we did not feed the Transformer with examples with prefix of size 0, as mentioned in Section \ref{models}, and we still want to have examples of only one decision-maker in each batch. We use the ADAM optimization algorithm (\cite{kingma2014adam}) with its default parameters as implemented in Pytorch: learning rate=$1e^{-03}$, fuzz factor $\epsilon=1e^{-08}$, and learning rate $decay$ over each update=$0.0$. We developed the DNNs with the AllenNLP software package (\cite{Gardner2017AllenNLP})\footnote{\url{https://allennlp.org/}} over Pytorch (\cite{paszke2017automatic}).

For our LSTM-based models we tune the LSTM hidden layer size (50, 80, 100, 200) and the number of LSTM layers (1, 2, 3). For our Transformer-based models, we tune the number of encoder and decoder layers (3, 4, 5, 6). In addition, we tune the dimension of the linear layers that follow the multi head attention layer of the encoder and decoder (0.5, 1, and 2 times of the input dimension), such that these parameters are the same for the encoder and the decoder. Finally, for the joint models, we tune the weight of the MSE, SCE and MSTPE losses (($\alpha, \beta, \gamma) \in ((1, 1, 1), (2, 2, 1), (1, 1, 2))$).

\section{Results}
\label{results}
As mentioned in Section \ref{experiment}, we perform two sets of comparisons in order to address the five research questions posed in Section \ref{task}, and in this section we will present their results. Both sets aim to answer the questions regarding our ability to predict the hotel choice rate and the per-trial decisions (questions \#1 and \#2, respectively), while comparing different modeling strategies (question \#3). The first set of comparisons focuses on the question that deals with text representations (question \#4), while the second set focuses on the complementary value of textual and behavioral representations (question \#5). 

\subsection{Per-Trial Prediction Results}
\label{per_trial_results}
Table \ref{tab:per_trial_results} presents the per-trial accuracy and macro average F1-score results for the various baselines and models, when using both the textual and the behavioral features. The table is divided to four sub-tables: The top table reports the results of our models when using the hand-crafted textual features, the next table reports the results of our models when using the BERT-based textual features, the next table reports the results of our models when using a concatenation of both BERT and our hand-crafted textual features, and the bottom table reports the results of the baselines.

The results show that the joint trial/choice-rate LSTM-TRCR model with our hand-crafted features is superior to the other models and baselines. 
%
The results are statistically significant in 3 folds when comparing LSTM-TRCR with Transformer-TRCR and Transformer-TR, and in 4 folds when comparing LSTM-TRCR to LSTM-TR (all these comparisons refer to models with our hand-crafted features). When comparing LSTM-TRCR with hand crafted features to any of the other models in the table, the resutls are significant in 5 or 6 folds.
%

The results also show that the Transformer-based models using BERT-based textual features, either alone or together with the hand-crafted features, do not perform well. 
In fact, they produce the same output as the MVC baseline. Note, however, that when the Transformer models are used with our hand-crafted features only, their performance substantially improves (e.g. the Transformer-TRCR model with our hand-crafted features is the second best model in terms of the macro average F1-score). These results provide a positive answer to our first research question by showing that our models, when using the hand-crafted textual features, perform better than the baselines according to both evaluation measures, especially in terms of the macro average F1-score. The differences between each of the models when using the hand-crafted features (the four models in the top block of the table) and the two baselines (bottom block of the table) are statistically significant in 5 folds, except for Transformer-TR, where the results are statistically significant in 2 folds.
%

We next show that the main reason for the superiority of our models is that they can predict, at least with a decent accuracy, the cases where the decision-maker deviates from its majority behavior (which is choosing the hotel). Table \ref{tab:per_trial_f_scores} presents the F1-scores of each of the classes: 'hotel' and 'stay at home'. Results are presented for the baselines and two of our best models: LSTM-TR and LSTM-TRCR with our hand-crafted features. It demonstrates that the LSTM-TRCR model performs almost as well as the baselines, when considering the F1-score of the 'hotel' class. It also shows that both models succeed in predicting some of the cases in which the decision-maker chooses to stay home, while the baselines perform poorly on this class (for the MVC baseline the score is 0 by definition, since this is not the majority decision in the training data).


\begin{table}[t]
\centering
\begin{tabular}{|l|c|c|}
\hline
\textbf{Model} & \textbf{\begin{tabular}[c]{@{}c@{}}Per-Trial \\ Accuracy\end{tabular}} & \textbf{\begin{tabular}[c]{@{}c@{}}Macro \\Average F1-score\end{tabular}} \\ \hline
\multicolumn{3}{|c|}{\textbf{Hand Crafted Features ($\mT_{HC}$)}} \\ \hline
\textbf{LSTM-TR}    &  76.5 & 70 \\ \hline
\textbf{LSTM-TRCR}   &  \textbf{76.9} & \textbf{70.5} \\ \hline
\textbf{Transformer-TR} & 75.3 & 60.1  \\ \hline
\textbf{Transformer-TRCR}  & 76.2 & 65.5 \\ \hline
\multicolumn{3}{|c|}{\textbf{BERT-based Features ($\mT_{DNN}$)}}   \\ \hline
\textbf{LSTM-TR}    &  61.3 & 51.8 \\ \hline
\textbf{LSTM-TRCR}   &  62.6 & 51.5  \\ \hline
\textbf{Transformer-TR} & 73.3 & 42.3  \\ \hline
\textbf{Transformer-TRCR}  & 73.3 & 42.3 \\ \hline
\multicolumn{3}{|c|}{\textbf{\begin{tabular}[c]{@{}c@{}}BERT-based Features +\\ Hand Crafted Features\end{tabular}}} \\ \hline
\textbf{LSTM-TR}    &  70.2 & 60  \\ \hline
\textbf{LSTM-TRCR}   &  71.8 & 61.2 \\ \hline
\textbf{Transformer-TR} &  73.3 & 42.3 \\ \hline
\textbf{Transformer-TRCR}  & 73.3 & 42.3 \\ \hline
\multicolumn{3}{|c|}{\textbf{Baselines}} \\ \hline
\textbf{MVC}     & 73.3 & 42.3 \\ \hline
\textbf{EWG}     & 59.8 & 50 \\ \hline
\end{tabular}
\caption{Per-trial performance with both textual and behavioral features.}
\label{tab:per_trial_results}
\end{table}

\begin{table}[h!]
\centering
\begin{tabular}{|l|c|c|}
\hline
\textbf{Model} & \textbf{\begin{tabular}[c]{@{}c@{}}F1-score \\ 'Hotel'\end{tabular}} & \textbf{\begin{tabular}[c]{@{}c@{}}F1-score \\ 'Stay Home'\end{tabular}} \\ \hline
\textbf{LSTM-TR}    &  83.8 & 56.1 \\ \hline
\textbf{LSTM-TRCR}   & 84.2 & \textbf{56.7} \\ \hline
\textbf{MVC}     & \textbf{84.6} & 0 \\ \hline
\textbf{EWG}     & 72.1 & 27.8 \\ \hline
\end{tabular}
\caption{Per-Class F1-scores for the per-trial analysis. LSTM-TR and LSTM-TRCR use the hand-crafted textual features and the behavioral features.}
\label{tab:per_trial_f_scores}
\end{table}

\subsection{Hotel Choice Rate Results}
\label{hotel_choice_rate_results}
Table \ref{proportion_results} presents our hotel choice rate results in terms of RMSE and bin macro average F1-score, when our models use both the textual and the behavioral features. The table is divided into four sub-tables, similarly to Table \ref{tab:per_trial_results}.

When considering the RMSE measure, the median (MED) and average (AVG) baselines, as well as the LSTM-CR model with our hand-crafted textual features, are superior, and the differences between the three models are not statistically significant. Also, the Transformer-CR and Transformer-TRCR models do not lag substantially behind, and the differences are statistically significant only between the AVG baseline and the Transformer-CR in all 6 folds. 
As discussed in Section \ref{Quantitative_Data_Analysis}, the behavior of most of our participants is similar to the average behavior in the entire group of participants. Hence, it is not surprising that the MED and AVG baselines excel under RMSE evaluation. 

In contrast, when considering the bin macro average F1-score, LSTM-TR with our hand-crafted textual features outperforms all other models and baselines, by a large margin (e.g. 48.3 compared to only 13.2 for the AVG baseline). 
The results are statistically significant in only 1 fold when comparing LSTM-TR with our hand-crafted textual features to Transformer-TR with either BERT or hand-crafted textual features. Further, they are statistically significant in 3 folds when comparing LSTM-TR with our hand-crafted textual features to the MED baseline, and in 4 to 6 folds for the other models and baselines. 
%
%
Generally, all our models but one substantially outperform the MVC, AVG and MED baselines on this measure, when using the hand-crafted features only. These results are statistically significant in at least 4 folds for all comparisons between a baseline and one of the DNN models except for 4 comparisons, in which the results are statistically significant in 2-3 folds. The same holds for the LSTM models when using the BERT features (with or without the hand-crafted features). These results provide a positive answer to our second research question by showing that our models can indeed learn to perform hotel choice rate prediction quite well. 




\begin{table}
\centering
\begin{tabular}{|l|c|c|}
\hline
\textbf{Model} & \textbf{RMSE}  & \textbf{\begin{tabular}[c]{@{}c@{}}Bin Macro \\Average F1-score\end{tabular}} \\ \hline
\multicolumn{3}{|c|}{\textbf{Hand Crafted Features ($\mT_{HC}$)}} \\ \hline
\textbf{SVM-CR} & 29.8 & 26.6 \\ \hline
\textbf{LSTM-TR} & 26.6 & \textbf{48.3} \\ \hline
\textbf{LSTM-CR} & \textbf{24.6} & 34.5 \\ \hline
\textbf{LSTM-TRCR} & 29.7 & 31.7 \\ \hline
\textbf{Transformer-TR} & 33.1 & 33.9  \\ \hline
\textbf{Transformer-CR} & 25.2 & 16.7 \\ \hline
\textbf{Transformer-TRCR} & 25.3 & 31.6 \\ \hline
\multicolumn{3}{|c|}{\textbf{BERT-based Features ($\mT_{DNN}$)}} \\ \hline
\textbf{SVM-CR} & 27.2 & 20.8 \\ \hline
\textbf{LSTM-TR} & 34 & 32.8 \\ \hline
\textbf{LSTM-CR} & 28.4 & 27.7 \\ \hline
\textbf{LSTM-TRCR} & 33.8 & 23.9 \\ \hline
\textbf{Transformer-TR} & 37.2 & 16.3  \\ \hline
\textbf{Transformer-CR} & 25.5 & 12.3 \\ \hline
\textbf{Transformer-TRCR} & 25.4 & 12.1 \\ \hline
\multicolumn{3}{|c|}{\textbf{BERT-based Features + Hand Crafted Features}} \\ \hline
\textbf{SVM-CR} & 27.4 & 21.9 \\ \hline
\textbf{LSTM-TR} & 30.4 & 39.2\\ \hline
\textbf{LSTM-CR} & 25.6 & 27.5\\ \hline
\textbf{LSTM-TRCR} & 28.3 & 26.2 \\ \hline
\textbf{Transformer-TR} & 37.2 & 16.3 \\ \hline
\textbf{Transformer-CR} & 25.5 & 12.8 \\ \hline
\textbf{Transformer-TRCR} & 25.4 & 12.3 \\ \hline
\multicolumn{3}{|c|}{\textbf{Baselines}} \\ \hline
\textbf{MVC} & 36.5 & 17.7 \\ \hline
\textbf{EWG} & 34.7 & 32.4 \\ \hline
\textbf{AVG} & \textbf{24.6} & 13.2 \\ \hline
\textbf{MED} & \textbf{24.6} & 16.2 \\ \hline
\end{tabular}
\caption{Hotel choice rate prediction performance of the baselines and the models with the textual and behavioral features. Note that for RMSE lower values are better, while for the bin macro average F1-score higher values are better.}
\label{proportion_results}
\end{table}

As in the per-trial prediction, the main reason for the superiority of our models is their ability to predict deviations from the majority behavior. Table \ref{tab:choice_rate_f_scores} presents the F1-scores of each of the bins defined in Section \ref{evaluation_measures}, both for the baselines and for the models with our hand-crafted textual features. It demonstrates that LSTM-TR outperforms all other models and baselines on two of the bins that relate to the non-majority behavior, while on the third non-majority bin Transformer-TR performs best and LSTM-TR is second best. The AVG baseline, in contrast, performs best on the majority behavior bin.\footnote{Although the MED baseline assigns to each test-set example the median hotel choice rate, as computed over the training set examples, it achieves a positive F1-score for two bins. This is because in our six fold cross-validation protocol, in two folds the median hotel choice rate of the training set examples falls within the third bin, while in the other four folds it falls within the fourth bin.}


\begin{table}[h!]
\centering
\begin{tabular}{|l|c|c|c|c|}
\hline
\textbf{Model} & \textbf{\begin{tabular}[c]{@{}c@{}}F1-score \\ Bin 1\end{tabular}} & \textbf{\begin{tabular}[c]{@{}c@{}}F1-score \\ Bin 2\end{tabular}} & \textbf{\begin{tabular}[c]{@{}c@{}}F1-score \\ Bin 3\end{tabular}} & \textbf{\begin{tabular}[c]{@{}c@{}}F1-score \\ Bin 4\end{tabular}} \\ \hline
\textbf{SVM-CR} & 21 & 18 & 44.6 & 23 \\ \hline
\textbf{LSTM-TR} & \textbf{50.9} & \textbf{33.1} & 47.9 & 61.4 \\ \hline
\textbf{LSTM-CR} & 26.4 & 15.3 & 42.8 & 53.5 \\ \hline
\textbf{LSTM-TRCR} & 26 & 15.6 & 26.7 & 58.4 \\ \hline
\textbf{Transformer-TR} & 24.5 & 21.1 & 24.5 & \textbf{65.5}  \\ \hline
\textbf{Transformer-CR} & 0 & 1.8 & 39.2 & 25.8 \\ \hline
\textbf{Transformer-TRCR} & 16.2 & 11.2 & 43 & 55.8 \\ \hline
\textbf{AVG} & 0 & 0 & \textbf{52.8} & 0 \\ \hline
\textbf{MED} & 0 & 0 & 17.6 & 47.1 \\ \hline
\end{tabular}
\caption{Hotel choice rate F1-scores for the four bins. The table presents the performance of the baselines and the models with the hand-crafted textual features and the behavioral features. The hotel choice rate mapping into the bins is: Bin 1: $choice \ rate < 0.25$, Bin 2: $0.25 \leq choice \ rate < 0.5$, Bin 3: $0.5 \leq choice \ rate < 0.75$, and Bin 4: $choice \ rate \geq 0.75$.}
\label{tab:choice_rate_f_scores}
\end{table}

Focusing on the results of the LSTM-based models, the choice rate model (LSTM-CR) outperforms the per-trial model (LSTM-TR) and the joint model (LSTM-TRCR) when considering the RMSE score, but the per-trial (LSTM-TR) model outperforms the other two models when considering the bin macro average F1-score. These patterns hold regardless of the type of textual features used by the models. These results and the results presented in Table \ref{tab:choice_rate_f_scores}, indicate that directly optimizing for the hotel choice rate is particularly useful for the overall RMSE performance. However, trial-based models better capture less frequent outcomes. Moreover, joint learning of both trial outcomes and the overall choice rate (with the LSTM-TRCR model) does not improve over learning only the trial-based outcome.
%


\subsection{The Impact of Different Textual Feature Sets}
\label{text_attribute_impact}

\begin{figure}[]
  \centering
  \includegraphics[width=0.48\textwidth]{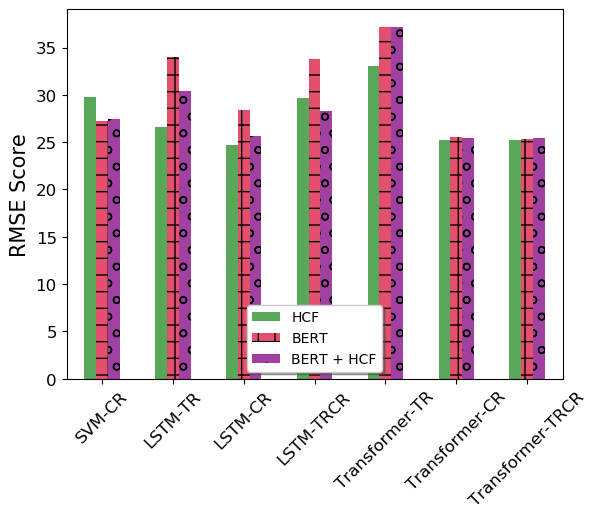}
 \includegraphics[width=0.48\textwidth]{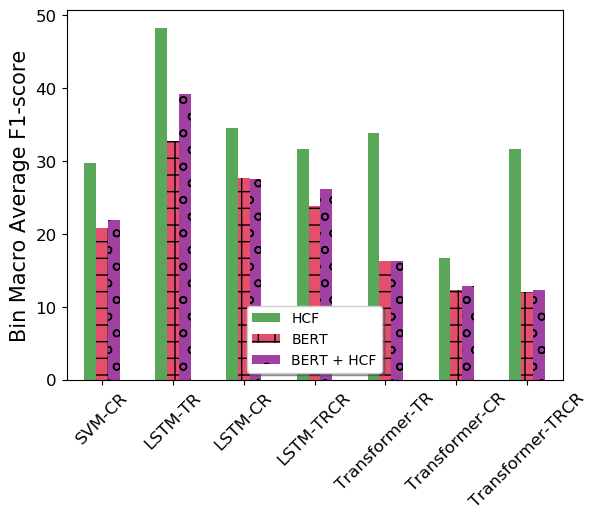}
  \caption{Textual features comparison for the hotel choice rate prediction task. The histogram presents the RMSE score (lower is better) and the bin macro average F1-score (higher is better) of each of our models, when using the behavioral features and each of our textual feature sets (HCF stands for 'hand-crafted features').}
 \label{bert_domain_features_compare_rmse_f_score}
\end{figure}

\begin{figure}[h!]
  \centering
  \includegraphics[width=0.48\textwidth]{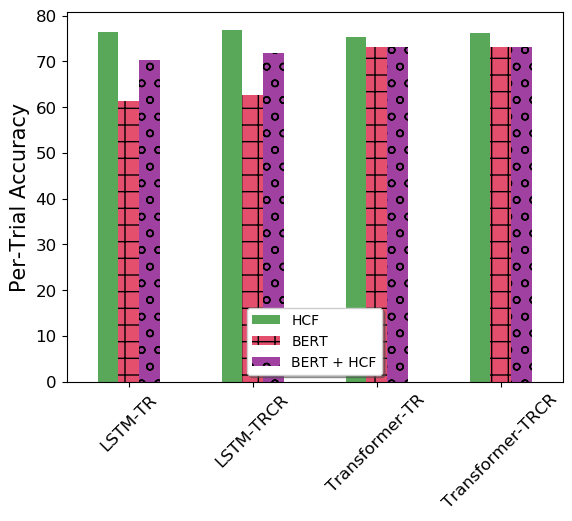}
  \includegraphics[width=0.48\textwidth]{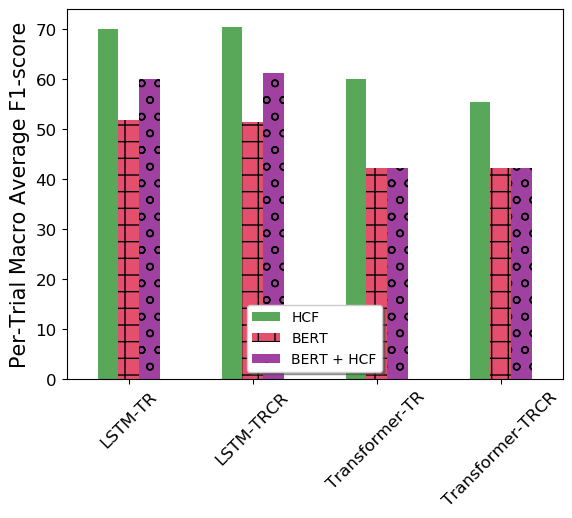}
  \caption{Textual features comparison for the per-trial prediction models. The histogram presents the per-trial accuracy (left) and the macro average F1-score (right) of each of our models, when using the behavioral features and each of our textual feature sets.}
 \label{bert_domain_features_compare_per_trial}
\end{figure}

Figures \ref{bert_domain_features_compare_rmse_f_score} and \ref{bert_domain_features_compare_per_trial} compare the performance of our models when using the different textual feature sets and the behavioral feature set. Figure \ref{bert_domain_features_compare_rmse_f_score} presents the hotel choice rate results, and it shows that except for SVM-CR, all our models achieve their best RMSE score (left part of the figure) when using our hand-crafted features. Likewise, the right part of Figure \ref{bert_domain_features_compare_rmse_f_score} indicates that all our models excel on the bin macro average F1-score when using the hand-crafted textual features, and the gaps are even larger. The results are statistically significant in at least 4 folds for LSTM-TR and both LSTM-TRCR, and Transformer-TRCR. Further, they are statistically significant in 3 folds for Transformer-CR and LSTM-CR. Finally, they are not statistically significant for Transformer-TR. Figure \ref{bert_domain_features_compare_per_trial} presents very similar patterns for the per-trial prediction models, and the results are statistically significant in at least 5 folds for all models. 
While the figures focus on models that use both the textual and the behavioral features, we observed very similar patterns when comparing models that use only the textual features. These results provide an answer to our fourth research question as they clearly indicate the value of our hand-crafted features, even when compared to a strong language representation model, like BERT.

\subsection{The Complementary Impact of Textual and Behavioral Features}

Table \ref{data_representation_results} presents the results of our set of comparisons which focuses on Question \#5, regarding the impact of the different feature sets: Behavioral and textual. As indicated above, in most cases our models perform best with our hand-crafted textual features. We hence focus on these textual features in the current comparison. 

Here, the table provides a mixed answer, where for some models and evaluation measures the joint feature set yields superior results, while for others it is better to use only the textual features. These results suggest that we should still deepen our understanding of the complementary effect of the two feature sets. This is a clear direction for future work. 

\begin{table}
\centering
\begin{tabular}{|c|c|c|c|c|c|}
\hline
\textbf{Model} & \textbf{Feature sets} & \textbf{RMSE} & \textbf{\begin{tabular}[c]{@{}c@{}}Bin \\ Macro\\ Average \\ F1-score\end{tabular}} & \textbf{\begin{tabular}[c]{@{}c@{}}Per-Trial\\ Accuracy\end{tabular}} & \textbf{\begin{tabular}[c]{@{}c@{}}Per-Trial \\ Macro \\ Average \\ F1-score\end{tabular}} \\ \hline \hline
SVM-CR & \begin{tabular}[c]{@{}c@{}}Only Textual\\Features\end{tabular} & \textbf{29.0} & \textbf{29.5} & - & - \\ \hline
SVM-CR & All Features & 29.8 & 26.6 & - & -\\ \hline \hline
LSTM-TR & \begin{tabular}[c]{@{}c@{}}Only Textual\\Features\end{tabular} & 27 & 47 & \textbf{77.6} & \textbf{71.2} \\ \hline
LSTM-TR & All Features & \textbf{26.6} & \textbf{48.3} & 76.5 & 70 \\ \hline \hline
LSTM-CR & \begin{tabular}[c]{@{}c@{}}Only Textual\\Features\end{tabular} & 26.6 & \textbf{34.7} & - & -\\ \hline
LSTM-CR & All Features & \textbf{24.7} & 34.5  & - & - \\ \hline \hline
LSTM-TRCR & \begin{tabular}[c]{@{}c@{}}Only Textual\\Features\end{tabular} & \textbf{25.9} & 27.7 & \textbf{78.8} & \textbf{72} \\ \hline
LSTM-TRCR & All Features & 29.7 & \textbf{31.7} & 76.9 & 70.5 \\ \hline \hline
Transformer-TR & \begin{tabular}[c]{@{}c@{}}Only Textual\\Features\end{tabular} & 33.2 & 31.7 & \textbf{75.5} & 57.7\\ \hline
Transformer-TR & All Features & \textbf{33.1} & \textbf{33.9} & 75.3 & \textbf{60.1}\\ \hline \hline
Transformer-CR & \begin{tabular}[c]{@{}c@{}}Only Textual\\Features\end{tabular} & \textbf{25.1} & \textbf{21.8} & - & -\\ \hline
Transformer-CR & All Features & 25.2 & 16.7  & - & - \\ \hline \hline
Transformer-TRCR & \begin{tabular}[c]{@{}c@{}}Only Textual\\Features\end{tabular} & \textbf{25.1} & \textbf{32.8} & 75.6 & 64.2\\ \hline
Transformer-TRCR & All Features & 25.3 & 31.6 & \textbf{76.2} & \textbf{65.5} \\ \hline
\end{tabular}
\caption{A comparison between all models with hand-crafted textual features (from both the prefix and suffix trials), behavioral features (from the prefix trials), as well as with a joint feature set. For each model and measure separately, we bold the feature set that outperforms the others.}
\label{data_representation_results}
\end{table}

\section{Which Linguistic Features are most Impactful - Interpretability Analysis}
\label{sec:interpret}

One of the main contributions of this paper is demonstrating the impact of verbal communication on decision making in our repeated persuasion game setup (Section \ref{sec:verbal}), and exploiting this signal in the prediction of the decision maker's decisions (Section \ref{results}). Moreover, our results demonstrate that our hand-crafted features are instrumental in achieving high quality predictions. This is an interesting finding since in most recent NLP work features that are automatically learned by DNNs (e.g. BERT in our case) provide a more effective text representation for NLP classifiers compared to hand-crafted features. 

Importantly, the superior performance of the hand-crafted features also paves the way for gaining an insight into which linguistic signals drive the choices of the decision maker. While such model interpretation analysis can also be applied to BERT-based models, most existing interpretation tools for such models would focus on the impact of individual words or simple textual phrases on the predictive power of the model. The hand-crafted features, in contrast, encode higher-level linguistic concepts (Section \ref{our_approach}) and hence allow us to gain a deeper understanding of the impact of various linguistic choices on persuasion (see discussion on word and phrase based vs. concept based interpretation of NLP models in \citep{feder2020causalm}).

We next aim to interpret the predictions of our best performing, hand-crafted-features based models,
so that we can gain such insights.


\subsection{Model Interpretation Methodology}

Machine learning model explanation and interpretation research has become dominant in recent years \citep{ws2019acl}. One particularly popular method is SHAP (SHapley Additive exPlanations) (\cite{lundberg2017unified}), providing a unified framework for interpreting the predictions of different models. This is a game theoretic approach which connects optimal credit allocation with local explanations using the classic Shapley values from game theory and their extensions. Although previous works made significant steps toward explaining DNN models by embracing the concept of Shapely value (e.g., \cite{DeepLIFT}), applications of this idea to the interpretation of sequential models such LSTM have proven much more challenging \citep{DeepLIFT,ho2021interpreting,dickinson2021positional}.


Hence, in order to shed light on the linguistic features that drive the prediction of the decision-makers' decisions, we follow a two-step process. First, we train a model that given the HC feature vector of each review, predicts the probability that a decision-maker chooses the Hotel option after observing the review. Specifically, we train three different tree-based boosting models (CatBoost, Random Forest and XGBoost),\footnote{We used the implementation of the scikit-learn API for these models: \url{https://catboost.ai/}, \url{https://scikit-learn.org/stable/modules/ensemble.html#forests-of-randomized-trees}, and \url{https://xgboost.readthedocs.io/en/latest/python/python_api.html#module-xgboost.sklearn}} following the same training protocol as in Section \ref{experiment} and select the best model based on the minimum RMSE criterion. We then use the SHAP framework\footnote{\url{https://shap.readthedocs.io/en/latest/index.html}} and rank the features based on their average SHAP values over the training set. 

While this analysis sheds light on the impact of certain HC features on the tendency of decision-makers to select the hotel described in the review, it does not address other important aspects of our setup such as its sequential, multi-stage nature. Hence, in the second step, we verify the impact of the top-ranked features by testing their ability to improve the performance of an LSTM-CR model with BERT-based features.  Particularly, we use 600 hotel reviews from the same Booking.com data set from which we sampled the reviews in our main experiment (see Section \ref{Quantitative_Data_Analysis}), to fine-tune a pre-trained BERT to predict the existence of the highest SHAP-rated features from the first step in each input review. We then utilize the resulting BERT model as the text representation layer of the LSTM-CR model (as described in Section \ref{models}). 

\begin{figure}[h]
  \centering
  \includegraphics[width=0.7\textwidth]{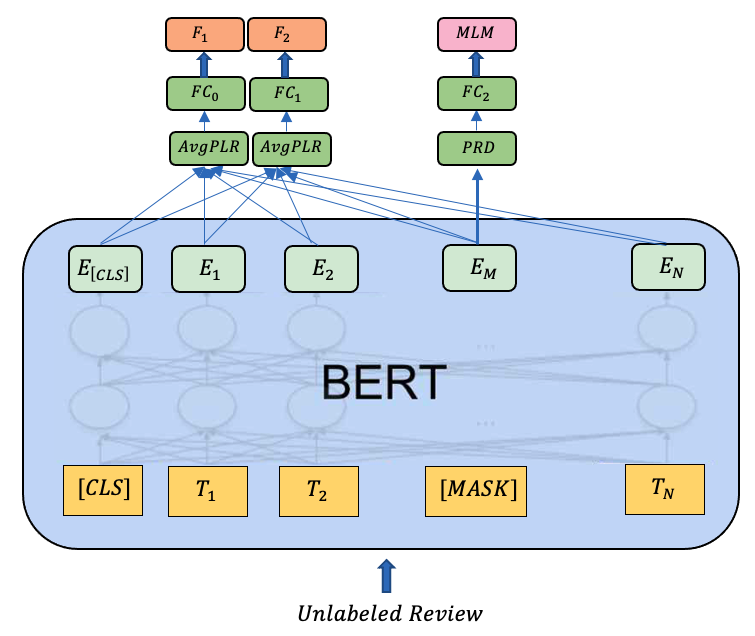}
  \caption{BERT fine-tuning with the prediction of two SHAP-based features. In this representative case, we add two tasks, $F_1$ and $F_2$, designed to predict whether these features are presented in the text. $PRD$ and $AvgPLR$ stand for the BERT prediction head and an average pooler head, respectively, $FC$ is a fully connected layer, and [MASK] stands for masked tokens embeddings.}
  \label{BERT_inter}
\end{figure}

For the BERT fine-tuning stage, our objective consists of the Masked Language Model (MLM) term along with auxiliary terms, designed to predict the existence of the $k \in \{1,2,3,4\}$ highest SHAP-rated HC features in the input review. With those tasks added to the loss function, we have that:

\[Loss = \frac{1}{n}(\sum\limits_{i=1}^{n}(L^i_{mlm}) + \sum\limits_{j=1}^{k}( \sum\limits_{i=1}^{n}(L^i_{feature_j}))\]
where $n$ is the number of samples, $L^i_{mlm}$ is the Cross Entropy Loss of the MLM task, and $L^i_{feature_j}$ is the Cross Entropy Loss of the $j$'th SHAP-based feature prediction task. Figure \ref{BERT_inter} illustrates this fine-tuning stage with two features.

To evaluate the importance of the top SHAP-ranked HC features, we construct another LSTM-CR model with BERT representations, this time fine-tuning BERT with $k \in \{1,2,3,4\}$ randomly selected HC features. We repeat this process five times, with different random samples of $k$ HC features, and report the average performance across these five runs.

To recap, our proposed method aims to identify impactful textual features using SHAP, despite the challenging nature of sequence model interpretation with this method. Our reasoning is that if the top SHAP-based ranked features for the (non-sequential) regression task of review persuasiveness also bias BERT in a direction that improves choice rate prediction when this BERT model is used as the textual representation layer of a LSTM-CR model, then these features are also of fundamental importance for persuasion in our setup. To the best of our knowledge, this interpretation framework for text-based sequential models is a novel contribution of this paper.


\subsection{Results}

Table \ref{shap_ranking} presents the top four SHAP-ranked HC textual features.  These features include: (a) word-based features such as positive sentiment words of the highest intensity as well as negative sentiment words with medium intensity; (b) a topical feature, namely, the discussion of the room in the negative part of the review; and (c) a more high-level semantic and structural concept, namely, an explicit statement, in the negative part of the review, that the author does not have any negative statement to make about the hotel. This set of impactful features demonstrates that review authors have in their disposal a variety of tools to persuade the decision makers. 

\begin{figure}[h!]
  \centering
  \includegraphics[width=0.7\textwidth]{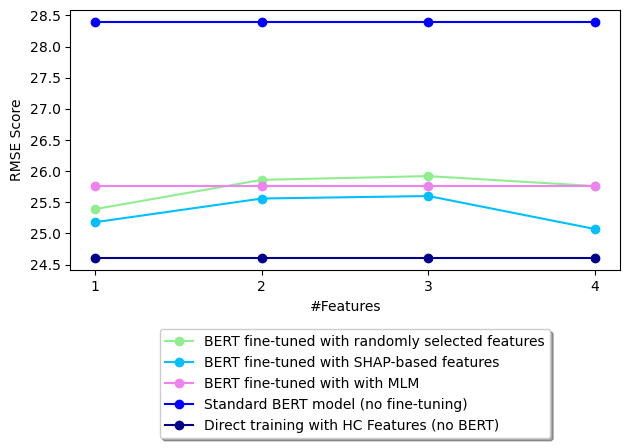}
  \caption{RMSE scores as a function of the number of features in the fine-tuning step of BERT. We compare BERT fine-tuning with features selected by SHAP, to BERT fine-tuning with randomly selected features. Results with the standard BERT (lower bound),  a BERT model fine-tuned with the standard Masked Language Model (MLM) objective on the same fine-tuning dataset (baselines), and of the LSTM-CR model when the texts are represented by the HC features (upper bound) are provided for reference.}
 \label{interpretability_section_results}
\end{figure}

As noted above, the top SHAP-ranked HC textual features are selected for a classifier that predicts the degree to which an individual review is persuasive. We hence also test the degree to which biasing the BERT encoder to encode information about these features improves the LSTM-CR model in our full setup. Figure \ref{interpretability_section_results} presents the results, using the RMSE measure. Clearly, for every value of $k \in \{1,2,3,4\}$ the LSTM-CR model performs better when BERT is fine-tuned with the SHAP-selected features, compared to when it is tuned with randomly selected features (tuning with randomly selected features actually leads to very similar performance as when just fine-tuning BERT on the same data with its standard Masked Language Modeling (MLM) objective). This analysis provides a complementary indication that the features highlighted above indeed have an important effect in the persuasion process in our setup.


\begin{table}[]
\begin{tabular}{|c|l|l|l|}
\hline
\textbf{Rank} & \textbf{High Level}                & \textbf{\begin{tabular}[c]{@{}l@{}}Feature Name\end{tabular}}      & \textbf{\begin{tabular}[c]{@{}l@{}}Feature Description\end{tabular}}                   \\ \hline
1     & \begin{tabular}[c]{@{}l@{}}Positive Part\\ Properties\end{tabular} & \begin{tabular}[c]{@{}l@{}}Words from the\\ third positive group\end{tabular}  & \begin{tabular}[c]{@{}l@{}}The positive part contains words\\ from the third positive group\end{tabular}         \\ \hline
2     & \begin{tabular}[c]{@{}l@{}}Negative Part\\ Properties\end{tabular} & Nothing negative                     & \begin{tabular}[c]{@{}l@{}}The negative part explicitly states\\ that there is nothing negative about\\ the hotel\end{tabular} \\ \hline
3     & \begin{tabular}[c]{@{}l@{}}Negative Part\\ Topics\end{tabular}   & Room                         & \begin{tabular}[c]{@{}l@{}}The negative part provides\\ details about the room\end{tabular}            \\ \hline
4     & \begin{tabular}[c]{@{}l@{}}Negative Part\\ Properties\end{tabular} & \begin{tabular}[c]{@{}l@{}}Words from the\\ second negative group\end{tabular} & \begin{tabular}[c]{@{}l@{}}The negative part contains\\ words from the second negative group\end{tabular}        \\ \hline
\end{tabular}
\caption{The top four SHAP-ranked HC textual features.  The positive and negative word groups are detailed in Table \ref{tab:semantic_group}. }
\label{shap_ranking}
\end{table}


We finally note that SHAP, as most model interpretation frameworks, is correlation-based. That is, it finds features that are correlated with the prediction of the model. Yet, correlation does not necessarily implies causation, and some of the highlighted features may in fact only be correlated with other, possibly unknown, features that truly drive persuasion. This discussion is, however, beyond the scope of this paper and we refer the interested reader to \citet{feder2020causalm} who present a causally-motivated model interpretation framework which aims to account for such phenomena.

\section{The Impact of Game Structure -- Ablation Analysis}
\label{ablation_analysis}

In Section \ref{results}, we addressed our five pre-defined research questions. In this section, we address additional aspects of our data, experiments, and results. In particular, we would like to discuss the quality of the models and baselines in predicting the labels of examples with various prefix sizes, and their quality in predicting the decisions in various stages of the interaction between the expert and the decision-maker.

\subsection{Prefix Size Impact}
\label{ablation_per_raisha}
In this part, we analyze the impact of the prefix size on our models. We are focusing on models that use behavioral features and hand-crafted textual features, that have demonstrated very effective in Section \ref{results}. If the prefix size affects the performance of our models, we may want to consider this parameter in the model selection process.

\begin{figure}[h!]
 \centering
 \includegraphics[width=0.48\textwidth]{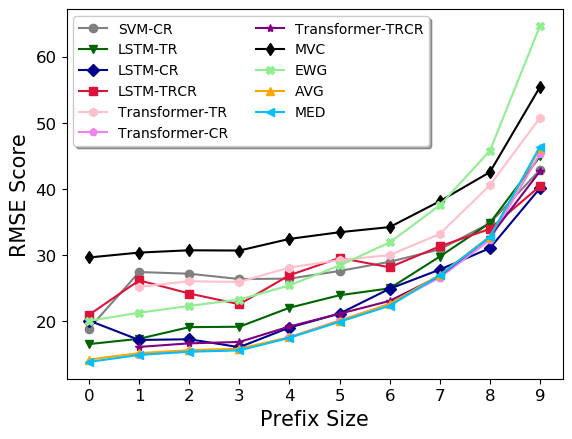}
  \includegraphics[width=0.48\textwidth]{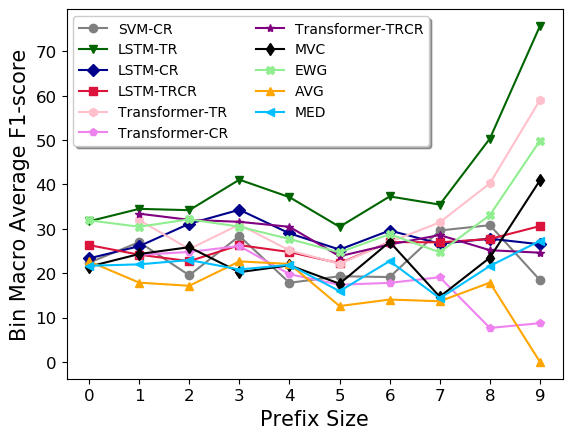}
 \caption{Prefix size impact on hotel choice rate performance. The graphs present the RMSE score (left) and the bin macro average F1-score (right) as a function of the prefix size.}
 \label{prefix_compare_RMSE}
\end{figure}

Figure \ref{prefix_compare_RMSE} presents the hotel choice rate performance, measured by RMSE (left) and bin macro average F1-score (right), as a function of the prefix size. The graphs indicate that while the RMSE score increases with the prefix size, for the bin macro average F1-score there is no strong correlation between the prefix size and the performance of the models, except for LSTM-TR, Transformer-TR, and the EWG and MVC baselines.

\begin{figure}[h!]
 \centering
 \includegraphics[width=0.48\textwidth]{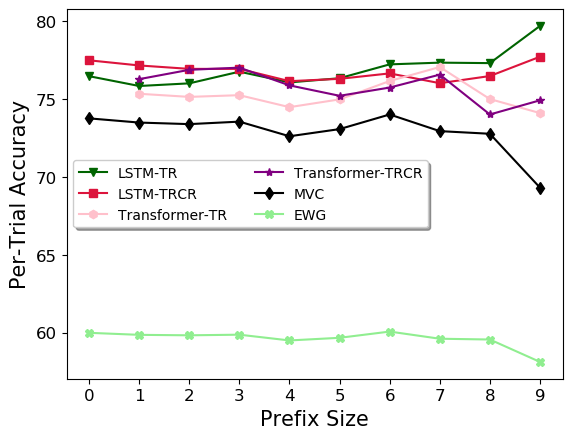}
  \includegraphics[width=0.48\textwidth]{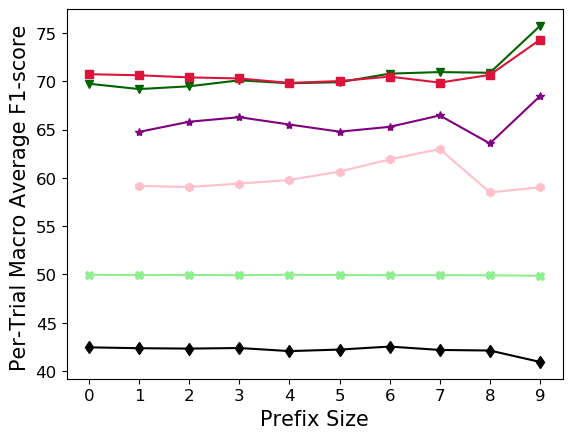}
 \caption{The effect of the prefix size on per-trail prediction performance. The graphs present the per-trial accuracy (left) and the macro average F1-score (right) as a function of the prefix size.}
 \label{per_trial_per_raisha_results}
\end{figure}

Figure \ref{per_trial_per_raisha_results} complements this hotel choice rate analysis and presents the per-trial accuracy (left) and the macro average F1 (right) of the per-trial and the joint models, as a function of the prefix size. The results show that the per-trial accuracy of the baselines somewhat decreases with the prefix size and are consistently worse than the performances of our models.
In contrast to the choice rate analysis, in this case there is no strong correlation between the prefix size and the performance of the models. 

The different correlations presented in these two figures, and particularly the difference between their left graphs, indicate that the prefix size has an impact when predicting the hotel choice rate, but not when predicting the decision at each trial. Understanding this difference is a potential important direction of future research. 

\subsection{Trial Number Impact}
\label{ablation_per_trial}
\begin{figure}[h!]
 \centering
 \includegraphics[width=0.48\textwidth]{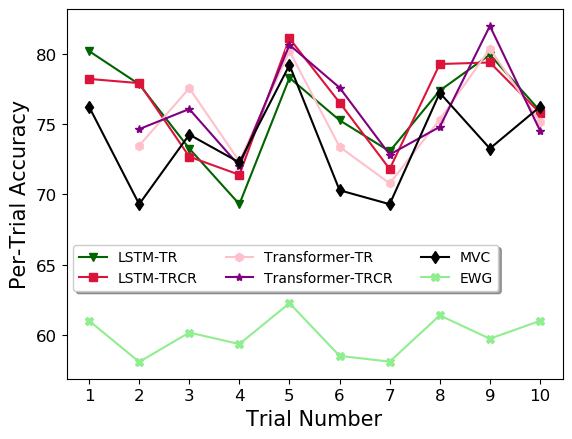}
  \includegraphics[width=0.48\textwidth]{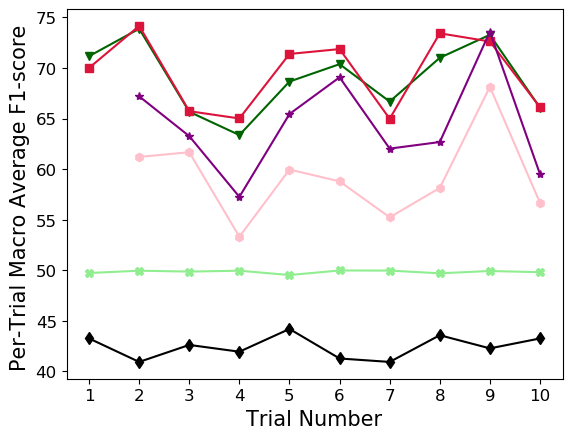}
 \caption{The trial number impact on the per-trial prediction results. The graphs present the per-trial accuracy (left) and the macro average F1-score (right) as a function of the trial number.}
 \label{per_trial_per_round_results}
\end{figure}

Figure \ref{per_trial_per_round_results} presents the per-trial accuracy (left) and the macro average F1-score (right) as a function of the trial number. Our motivation for this analysis is exploring the temporal dynamic between the expert and the decision-maker and the development of their mutual trust. In Section \ref{data}, we show that the hotel choice rate changes as the experiment progresses, but the changes are very small. We also show that the decision in each trial depends on the decision and the feedback in the previous trial. Here we explore whether temporal patterns can also be observed in the predictions of our models.

The figures demonstrate a temporal dynamic in the predictions of our models, although not necessarily an expected one. Particularly, the performance seem to have a periodical behavior such that after a maximum or a minimum point is achieved, the performance starts to move in the opposite direction. Since this is also the pattern we observe for the MVC baseline, these results seem to be explained by the pattern of human deviation from the majority behavior. Explaining this pattern is another interesting direction of future work.


\section{Discussion}
\label{discussion}

We explored the task of predicting the decisions made in language-based repeated persuasion games. For this purpose, we conducted an online experiment with more than 500 pairs of participants. In contrast to previous behavioral economics work where the experts and decision-makers communicate through numerical messages, the messages are verbal in our setting.
We explored five research questions, divided to sets (Section \ref{Introduction}): One about the nature of the prediction task (first three questions) and the other about the importance of the linguistic signal to our setup (last two questions). We will next summarize the main findings and conclusions related to these questions.

The \textbf{first} question focuses on our ability to predict the decisions in each trial given the history of $pr$ trials, where $pr$ is an integer number between 0 and 9.
We demonstrated that DNN modeling combined with our hand-crafted textual features as well as behavioral features (LSTM-TR) is the superior modeling strategy. Interestingly, when considering the per-trial accuracy evaluation measure, the MVC baseline is comparable to our strongest models. In contrast, our models are superior when the evaluation is done with the macro average F1-score. This is mostly because the baselines fail to detect the less dominant class in our data - decisions not to choose the hotel. This failure makes them poor prediction models that only excel in detecting the dominant class in the data, while our best models can go beyond that and also predict minority classes.

Our \textbf{second} research question targets our models' ability to accurately predict the hotel choice rate in a given suffix of a communication sequence. More than half of the participants select the hotel choice in seven or eight trials, and the average hotel choice rate was 7.18, with a standard deviation of 1.52. Therefore, predicting the average or median hotel choice rate, in our case, is a good approximation of the behavior of most participants. Indeed, in terms of the RMSE score, the simple MED and AVG baselines were the superior modeling strategies (together with LSTM-CR). However, unlike our models, these baselines fail to predict decisions that deviate from our population's average behavior. This is reflected by the superior bin macro average F1-score performance of our models. As in our findings for the first question, we observe that our models are valuable for predicting deviations from the population's most frequent behaviors.

Our \textbf{third} research question aims to identify the ideal modeling strategy for our setup. We explored three strategies: a non-structured classifier, a sequence model, and an attention-based approach.
We show that in the per-trial prediction task, LSTM-TRCR outperforms the other models and baselines. We also show that in the hotel choice rate prediction task, LSTM-CR outperforms the other models in terms of the RMSE score, and LSTM-TR outperforms the other models and baselines in terms of the bin macro average F1-score. These results indicate that a sequence model is the ideal modeling strategy for our setup. We hypothesize that the attention-based models are inferior to the sequence models in our setup, due to the relatively small training data set which is particularly challenging for highly parameterized models such as the Transformer. Moreover, we show that the SVM-CR non-structured classifier is consistently inferior to the other models, emphasizing the need for DNN modeling in our setup.

We have further considered models (both sequential and attention-based) that jointly consider the hotel choice rate and the per-trial decisions. These models achieved good results in both tasks and outperformed the other models and baselines in the per-trial prediction. 

As noted above, the second set of research questions considers the importance of the linguistic signal to our setup. In Section \ref{sec:verbal} we have established that our subjects behave differently in the presence of linguistic signals, compared to when signaling is numerical in nature, and that this behavior increases the descriptive value of a particular equilibrium (a full pooling equilibrium). This motivates us to also explore the impact of the textual messages and the linguistic features that represent them in our prediction models, as done in our second set of research questions.

The \textbf{fourth} question is designed to find the textual features that most serve our prediction model. We compared two sets of textual features: hand-crafted and DNN-based (BERT). We show that all our models, except for SVM-CR, achieved their best results when using the hand-crafted features, regardless of the evaluation measure used. This may be an indication that BERT is not capable of capturing the high-level task-related semantics of the task, as encoded in our hand-crafted features.

Finally, our \textbf{fifth} question focuses on the different aspects of the data which are crucial for our prediction. Particularly, we explored the impact of the textual messages, and the complementary effect of both the textual messages and the decision-makers' behavior throughout the game. Interestingly, our results indicate a mixed answer to this question, with different models performing best with different feature sets. Hence, we cannot provide a conclusive answer to this question.

In addition to these research questions, in Section \ref{sec:interpret} we also explored which textual features are most influential in our prediction process. We have proposed a novel model interpretation process, that aims to overcome the limitations of SHAP in dealing with sequential models. Our analysis reveals word-based, topical and structural concepts that drive the persuasion prediction capability of our LSTM-CR model. Gaining in-depth insight into the linguistic signals that drive persuasion is a prominent direction of future research, and our work has established its importance and proposed an initial algorithmic approach.

In this work, we have chosen to focus on predicting the decision-maker's decisions. These decisions partially determine the payoff of both the expert and the decision-maker, although the latter's payoff also depends on a random coin flip. Naturally, in the scope of one paper we could not focus on other important tasks, such as predicting the expert's decisions, or on predictions that would give in-depth insights about our setup, such as predicting the hotel choice rate as a function of the hotel's expected value. We leave these as well as other important questions for future work, hoping that our novel task and data set will pave the way for other researchers to explore the use of natural language in persuasion games.

One challenging aspect of our work is the generalization to new hotels. As described in Section \ref{data_collection}, we used one set of hotels in the train and development set and another set of hotels in the test set. Therefore, our models should generalize across different reviews in order to perform well on the test set. 
We also performed experiments where the test set has the same set of hotels and reviews as the training and development set. In the hotel choice rate prediction task, the results show that the SVM-CR model, when using the hand-crafted textual features and the behavioral features, outperforms all other models and baselines in terms of both measures. The results also show that LSTM-CR and LSTM-TRCR, when using the hand-crafted textual features and the behavioral features, outperform the baselines on both measures. This is in contrast to the main results of this paper, where cross-hotel generalization is required and the AVG and MED baselines achieve the best choice rate RMSE score (together with LSTM-CR), while LSTM-TR performs best in the choice rate bin analysis. For the per-trial decision task, the results are similar to those we show in Section \ref{results} in terms of the relative ranking of the models and the baselines, with slightly better performance of our models and slightly worse performance of the baselines. This comparison shows that the LSTM-based models excel in both conditions, indicating their robustness.

Another particular challenging aspect of our work, which may be considered a limitation, is the use of lab data, as opposed to real-world interactions. On the one hand, the use of lab data lets us control the various aspects of our experiments and allow us to draw intricate conclusions. On the other hand, previous studies revealed an interesting gap when comparing lab and field studies of social interactions (see review in \cite{levitt2007laboratory} and one demonstration of this gap in \cite{gneezy2004inefficiency}). Therefore, one of our main directions for future work is to explore whether our results generalize to real-world setups.

We would finally like to highlight two future work directions that seem of particular interest to us. The first direction has to do with online learning. While we predict the future behavior of the decision maker under a pre-defined policy of the expert (i.e. a pre-defined set of reviews chosen to describe the hotels in the suffix), in online learning the policy of both agents may change as a function of the input they receive from the world (e.g. the random hotel score (feedback) observed by both agents after the decision-maker makes her decision). In contrast to our task that focuses on pre-defined policy evaluation, online learning provides more dynamic policy adjustment capabilities. The second direction goes even further and utilizes our data in order to learn automatic agents that can simulate either the expert or the decision maker, performing their task optimally under a pre-defined criterion (e.g. maximizing the agent's gains or, alternatively, the mutual benefits of both players). Moving from behavior prediction to generation reflects a deeper understanding of our task and data. As noted above, we hope that our task, data and algorithms will draw the attention of the research community to language-based persuasion games and will facilitate further research on these questions as well as others.

\section*{Acknowledgements}
The work of R, Apel and M. Tennenholtz is funded by the European Research Council (ERC) under the European Union?s Horizon 2020 research and innovation programme (grant agreement n$\degree$  740435).

\clearpage
\appendix

\section{Experimental details}

\subsection{Task Description}
\label{appendix:screen}

\begin{figure}[h!]
  \centering
  \includegraphics[scale=0.75,valign=t]{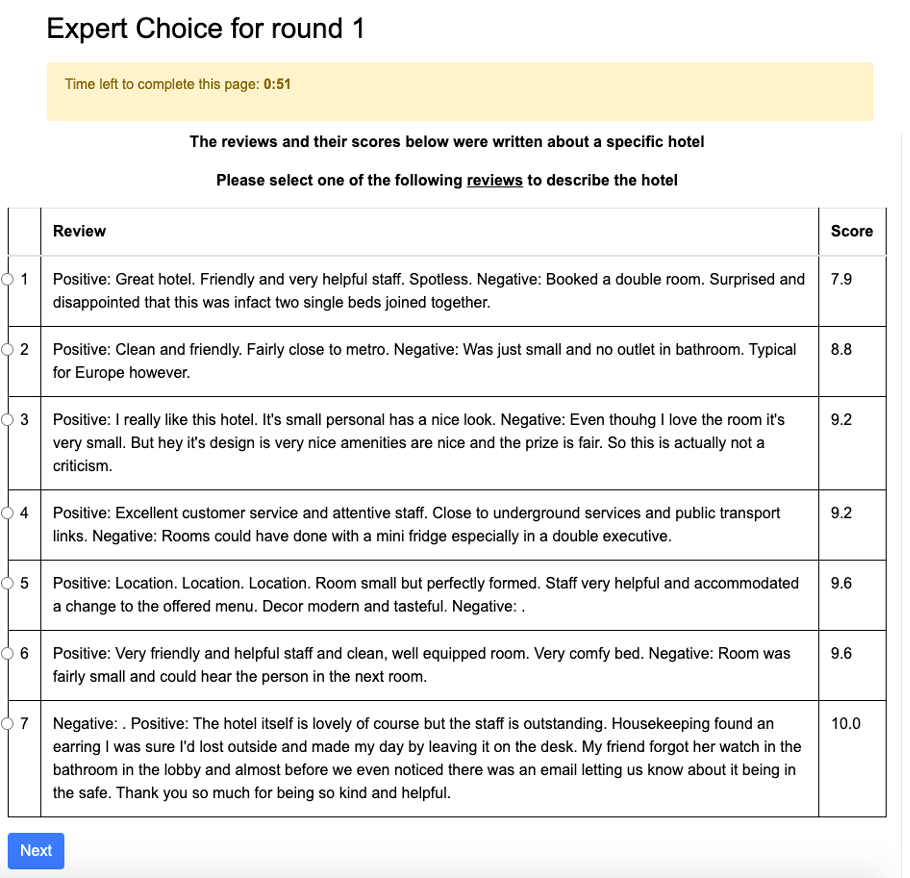}
  \includegraphics[scale=0.85,valign=t]{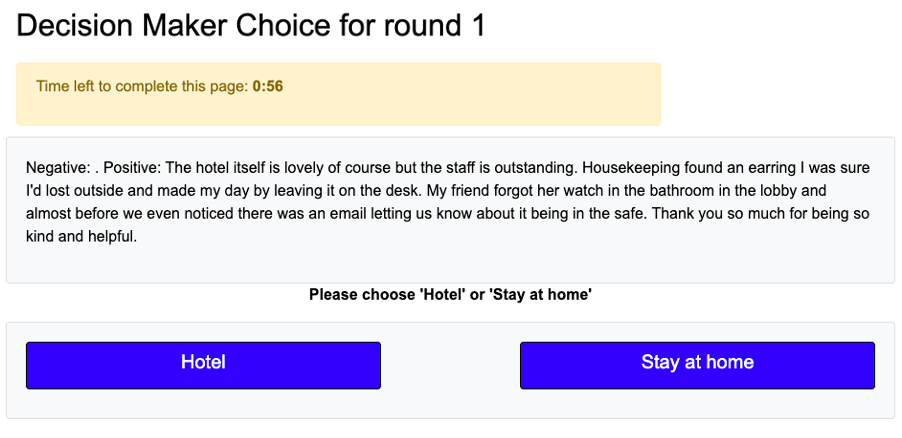}
  \caption{Screen-shots of the tasks presented to the decision-maker and to the expert.}
 \label{exp_screen_shots}
\end{figure}

\clearpage

\subsection{Experiment Instructions (Verbal condition)}
\label{appendix:instructions}

\subsubsection{Expert Instructions}

\textbf{You have 10 minutes to finish reading the instructions! If you are not to meet this deadline, the HIT will be automatically submitted, and you will not get paid.} \newline
You are invited to participate in an experiment. The experiment is part of a research conducted in the Industrial Engineering and Management Faculty of the Technion, Haifa. \newline
You have been randomly and anonymously paired with another participant. You were randomly selected to be the \textbf{Expert} (a travel agent), while your partner is the Decision Maker (your client). \newline
The experiment consists of 10 trials, played one after the other. All trials will be played with the same partner, and each of you will have the same role during the whole experiment. \newline
In each trial, you will be asked to sell to your client (the Decision Maker) a different hotel, and s/he will be asked to choose between the ‘Hotel’ (that provides a gain or a loss), and ‘stay at home’ (payoff 0 with certainty). \newline
The Decision Maker will not have prior information about the hotel.
You will get a sample of 7 reviews that were written by previous visitors to the hotel and their scores. You will be asked to choose one of those reviews and \textbf{only this review} will be shown to your partner as an estimation of the hotel’s quality. \newline
Your payoff in each trial will be a gain of 1 point if the Decision Maker chose Hotel, and 0 points otherwise. \newline
The Decision Maker's payoff will be as following:
If s/he chooses the 'Hotel' option, her/his payoff will be the benefit from the hotel minus a fixed cost of 8 points. After each of you has made your choices, the benefit will be chosen randomly from the 7 given scores. All scores are in the range between 2.5 and 10. \newline
If the Decision Maker chooses the 'Stay at Home' option, his/her payoff will be 0. \newline
In the comment text box below, please type “sdkot” (without commas and quotes), so we can be sure you are reading this. If you fail to do so, you will be unable to complete this HIT. You have to be quick. You will have 240 seconds to respond in the first 2 trials, and 240 seconds for the subsequent trials. If you are not to meet these deadlines, the Decision Maker will receive the review with the lowest score, and your payoff for the trial will be 0. \newline
\textbf{Risks and benefits:} \newline
There is no risk. Your benefit is contributing to a research project. \newline
\textbf{Compensation:} \newline
You will get 2.5 for your participation. At the end of the experiment, the accumulated points will set your probability to get a bonus of \$1.
Your goal is to maximize your earnings. \newline
\textbf{Voluntary participation and right to withdraw:} \newline
Participation in this study is voluntary, and you can stop at any time without any penalty. To finish, click on the “Return HIT” button or close your browser window. \newline
\textbf{Contact information:} \newline
If you have any concerns or questions about this research, you may contact: Reut Apel at reutapel@campus.technion.ac.il \newline
Have Fun! \newline
Do you have any comments on this HIT? \newline

\textbf{Clicking accept} \newline
By clicking on the “I agree” button, you indicate that you voluntarily agree to participate in this experiment and that you understand the information in this form.

\subsubsection{Decision-Maker Instructions}

\textbf{You have 10 minutes to finish reading the instructions! If you are not to meet this deadline, the HIT will be automatically submitted, and you will not get paid.} \newline
You are invited to participate in an experiment. The experiment is part of a research conducted in the Industrial Engineering and Management Faculty of the Technion, Haifa. \newline
\textbf{Pay attention:} at the end of the experiment, you will be asked to answer one simple question to verify you indeed read the texts in the experiment. If you do not answer it correctly, we will reject your work, and you will not get paid. \newline
You have been randomly and anonymously paired with another participant. You were randomly selected to be the \textbf{Decision Maker} (a traveler), while your partner is the Expert (your travel agent). \newline
The experiment consists of 10 trials, played one after the other. All trials will be played with the same partner, and each of you will have the same role during the whole experiment. \newline
In each trial, the Expert will try to sell you a different hotel, and you will be asked to choose between the ‘Hotel’ (that provide a gain or a loss), and ‘stay at home’ (payoff 0 with certainty). \newline
You will not have prior information about the hotel. \newline
The Expert will have some information and will send you a verbal estimation of the hotel’s quality. \newline
The choice will be made by clicking on one of the buttons: ‘Hotel’ or ‘Stay at home’. \newline
The Expert will gain of 1 point if you choose Hotel, and 0 points otherwise. \newline
Your payoff will be as following: \newline
If you choose the 'Hotel' option, your payoff will be the benefit from the hotel minus a fixed cost of 8 points. After each of you has made your choices, the benefit will be chosen randomly from the scores given by previous visitors to this hotel. All scores are in the range between 2.5 and 10. \newline
If you choose the 'Stay at Home' option, your payoff will be 0. \newline
In the comment text box below, please type “sdkot” (without commas and quotes), so we can be sure you are reading this. If you fail to do so, you will be unable to complete this HIT. You have to be quick. You will have 60 seconds to respond in the first 2 trials, and 40 seconds for the subsequent trials. If you are not to meet these deadlines, one of the options will be chosen randomly. If the Expert is too slow, you will receive a default evaluation. \newline
\textbf{Please be patient;} the Expert's role requires more time. Therefore, there will be a gap between one trial to another. \newline
\textbf{Risks and benefits:} \newline
There is no risk. Your benefit is contributing to a research project. \newline
\textbf{Compensation:} \newline
You will get 2.5 for your participation. At the end of the experiment, the accumulated points will set your probability to get a bonus of \$1.
Your goal is to maximize your earnings. 
You will get 18.2 points at the beginning of the experiment. \newline
\textbf{Voluntary participation and right to withdraw:} \newline
Participation in this study is voluntary, and you can stop at any time without any penalty. To finish, click on the “Return HIT” button or close your browser window. \newline
\textbf{Contact information:} \newline
If you have any concerns or questions about this research, you may contact: Reut Apel at reutapel@campus.technion.ac.il \newline
Have Fun! \newline
Do you have any comments on this HIT? \newline

\textbf{Clicking accept}
By clicking on the “I agree” button, you indicate that you voluntarily agree to participate in this experiment and that you understand the information in this form.

\clearpage

\section{Data}
\label{appendix:data}

\begin{table}[h!]
\centering
\begin{tabular}{ccccc}
\hline
\multicolumn{5}{|c|}{\textbf{Train-Validation hotels}}\\ \hline
\multicolumn{1}{|c|}{\textbf{\#hotel}} & \multicolumn{1}{c|}{\textbf{\begin{tabular}[c]{@{}c@{}}Average\\ score\end{tabular}}} & \multicolumn{1}{c|}{\textbf{\begin{tabular}[c]{@{}c@{}}Median\\ score\end{tabular}}} & \multicolumn{1}{c|}{\textbf{\begin{tabular}[c]{@{}c@{}}Scores’\\ STD\end{tabular}}} & \multicolumn{1}{c|}{\textbf{\begin{tabular}[c]{@{}c@{}}Hotel’s scores \\ distribution\end{tabular}}} \\ \hline
\multicolumn{1}{|c|}{1} & \multicolumn{1}{c|}{4.17} & \multicolumn{1}{c|}{3.8} & \multicolumn{1}{c|}{1.39} & \multicolumn{1}{c|}{2.5, 3.3, 3.3, 3.8, 4.2, 5.8, 6.3} \\ \hline
\multicolumn{1}{|c|}{2} & \multicolumn{1}{c|}{6.66} & \multicolumn{1}{c|}{7.9} & \multicolumn{1}{c|}{2.45} & \multicolumn{1}{c|}{3.3, 3.3, 6.3, 7.9, 8.3, 8.3, 9.2} \\ \hline
\multicolumn{1}{|c|}{3} & \multicolumn{1}{c|}{7.44} & \multicolumn{1}{c|}{7.5} & \multicolumn{1}{c|}{1.63} & \multicolumn{1}{c|}{5, 5.8, 7.1, 7.5, 8.3, 8.8, 9.6}\\ \hline
\multicolumn{1}{|c|}{4} & \multicolumn{1}{c|}{7.97} & \multicolumn{1}{c|}{8.3} & \multicolumn{1}{c|}{1.91} & \multicolumn{1}{c|}{5.4, 5.8, 7.1, 8.3, 9.2, 10, 10} \\ \hline
\multicolumn{1}{|c|}{5} & \multicolumn{1}{c|}{8.11} & \multicolumn{1}{c|}{8.8} & \multicolumn{1}{c|}{1.62} & \multicolumn{1}{c|}{5.8,6.3, 7.5, 8.8, 8.8, 9.6, 10} \\ \hline
\multicolumn{1}{|c|}{6} & \multicolumn{1}{c|}{8.33} & \multicolumn{1}{c|}{8.3} & \multicolumn{1}{c|}{1.79} & \multicolumn{1}{c|}{5.4, 7.1, 7.5, 8.3, 10, 10, 10} \\ \hline
\multicolumn{1}{|c|}{7} & \multicolumn{1}{c|}{8.94} & \multicolumn{1}{c|}{9.2} & \multicolumn{1}{c|}{1.34} & \multicolumn{1}{c|}{6.3, 8.3, 8.8, 9.2, 10, 10, 10} \\ \hline
\multicolumn{1}{|c|}{8} & \multicolumn{1}{c|}{9.19} & \multicolumn{1}{c|}{9.2} & \multicolumn{1}{c|}{0.68} & \multicolumn{1}{c|}{7.9, 8.8, 9.2, 9.2, 9.6, 9.6, 10} \\ \hline
\multicolumn{1}{|c|}{9} & \multicolumn{1}{c|}{9.54} & \multicolumn{1}{c|}{9.6} & \multicolumn{1}{c|}{0.28} & \multicolumn{1}{c|}{9.2, 9.2, 9.6, 9.6, 9.6, 9.6, 10} \\ \hline
\multicolumn{1}{|c|}{10} & \multicolumn{1}{c|}{9.77} & \multicolumn{1}{c|}{9.6} & \multicolumn{1}{c|}{0.21} & \multicolumn{1}{c|}{9.6, 9.6, 9.6, 9.6, 10, 10, 10} \\ \hline
\multicolumn{2}{|c|}{\textbf{Average Score}} & \multicolumn{3}{c|}{\textbf{8.01}} \\ \hline
\multicolumn{1}{l}{} & \multicolumn{1}{l}{} & \multicolumn{1}{l}{} & \multicolumn{1}{l}{} & \multicolumn{1}{l}{} \\ \hline
\multicolumn{5}{|c|}{\textbf{Test hotels}} \\ \hline
\multicolumn{1}{|c|}{\textbf{\#hotel}} & \multicolumn{1}{c|}{\textbf{\begin{tabular}[c]{@{}c@{}}Average\\ score\end{tabular}}} & \multicolumn{1}{c|}{\textbf{\begin{tabular}[c]{@{}c@{}}Median\\ score\end{tabular}}} & \multicolumn{1}{c|}{\textbf{\begin{tabular}[c]{@{}c@{}}Scores’\\ STD\end{tabular}}} & \multicolumn{1}{c|}{\textbf{\begin{tabular}[c]{@{}c@{}}Hotel’s scores \\ distribution\end{tabular}}} \\ \hline

\multicolumn{1}{|c|}{1} & \multicolumn{1}{c|}{4.56} & \multicolumn{1}{c|}{4.2} & \multicolumn{1}{c|}{0.97} & \multicolumn{1}{c|}{3.8, 3.8, 3.8, 4.2, 4.6, 5.4, 6.3} \\ \hline
\multicolumn{1}{|c|}{2} & \multicolumn{1}{c|}{6.69} & \multicolumn{1}{c|}{6.7} & \multicolumn{1}{c|}{2.09} & \multicolumn{1}{c|}{4.2, 4.2, 6.3, 6.7, 7.1, 8.3, 10} \\ \hline
\multicolumn{1}{|c|}{3} & \multicolumn{1}{c|}{7.74} & \multicolumn{1}{c|}{8.3} & \multicolumn{1}{c|}{1.78} & \multicolumn{1}{c|}{5, 5.8, 7.5, 8.3, 8.8, 8.8, 10}\\ \hline
\multicolumn{1}{|c|}{4} & \multicolumn{1}{c|}{7.91} & \multicolumn{1}{c|}{7.9} & \multicolumn{1}{c|}{1.49} & \multicolumn{1}{c|}{5.8, 6.7, 7.5, 7.9, 7.9, 9.6, 10}\\ \hline
\multicolumn{1}{|c|}{5} & \multicolumn{1}{c|}{8.04} & \multicolumn{1}{c|}{8.3} & \multicolumn{1}{c|}{1.24} & \multicolumn{1}{c|}{5.8, 7.5, 7.5, 8.3, 8.8, 8.8, 9.6} \\ \hline
\multicolumn{1}{|c|}{6} & \multicolumn{1}{c|}{8.33} & \multicolumn{1}{c|}{8.3} & \multicolumn{1}{c|}{1.82} & \multicolumn{1}{c|}{5, 7.1, 8.3, 8.3, 9.6, 10, 10} \\ \hline
\multicolumn{1}{|c|}{7} & \multicolumn{1}{c|}{8.89} & \multicolumn{1}{c|}{8.8} & \multicolumn{1}{c|}{1.34} & \multicolumn{1}{c|}{6.3, 8.3, 8.8, 8.8, 10, 10, 10} \\ \hline
\multicolumn{1}{|c|}{8} & \multicolumn{1}{c|}{9.04} & \multicolumn{1}{c|}{9.2} & \multicolumn{1}{c|}{0.99} & \multicolumn{1}{c|}{7.9, 7.9, 8.3, 9.2, 10, 10, 10} \\ \hline
\multicolumn{1}{|c|}{9} & \multicolumn{1}{c|}{9.65} & \multicolumn{1}{c|}{9.6} & \multicolumn{1}{c|}{0.42} & \multicolumn{1}{c|}{8.8, 9.6, 9.6, 9.6, 10, 10, 10}\\ \hline
\multicolumn{1}{|c|}{10} & \multicolumn{1}{c|}{9.66} & \multicolumn{1}{c|}{9.6} & \multicolumn{1}{c|}{0.43} & \multicolumn{1}{c|}{8.8, 9.6, 9.6, 9.6, 10, 10, 10} \\ \hline
\multicolumn{2}{|c|}{\textbf{Average Score}} & \multicolumn{3}{c|}{\textbf{8.06}} \\ \hline
\end{tabular}
\centering
\caption{Hotels' score distributions in our train-validation and test sets.}
  \label{hotels_distributions}
\end{table}

\clearpage

\begin{table}[h!]
\centering
\begin{tabular}{|l|l|l|l|}
\hline
  & \textbf{Review} & \textbf{Score} & 
  \begin{tabular}[c]{@{}l@{}} \textbf{hotel's} \\ \textbf{scores} \\ \textbf{distribution} \end{tabular} \\ \hline
1 & \begin{tabular}[c]{@{}l@{}}\textbf{Negative:} The swimming pool on the top of the roof is \\ very small. In high season there is little possibility that \\ you will be able to use it. The worst thing was that \\ during my stay the crew started to paint all \\ the walls on my floor's corridor. \\ The paint smell really awful. Although the stuff from the \\ Reception desk was ok the women bartender who worked \\ on morning shift wasn't very nice maybe she felt a little \\ bit sleepy. In my opinion the cost was too high compared \\ to the offer. \textbf{Positive:} The location is awesome. \\ You can go across the street and grab a subway. \\ The Sagrada Familia is about 15 20 minutes by foot.\end{tabular} & 5.8   & 
\begin{tabular}[c]{@{}l@{}} 5.8, 6.3, 7.5, \\ 8.8, 8.8, 9.6, \\ 10 \end{tabular}     \\ \hline
2 & \begin{tabular}[c]{@{}l@{}}\textbf{Positive}: The whole experience of our trip to Barcelona \\ and the hotel was perfect. \\ I can not speak highly enough of everyone who made our \\ stay so special. Our room was lovely and clean with \\ a fantastic shower and huge comfy bed. \\ We spent time in the spa and on the roof terrace which \\ has spectacular views over the city very close to the metro \\ so getting about was easy I will return \\ here I hope sometime in the future. \textbf{Negative:} \\ I really cannot think of anything at the moment.\end{tabular} & 10  & 
\begin{tabular}[c]{@{}l@{}} 5.8, 6.3, 7.5, \\ 8.8, 8.8, 9.6, \\ 10 \end{tabular}    \\ \hline
3 & \begin{tabular}[c]{@{}l@{}}\textbf{Negative:} 1. we didn't received what we asked a room \\ with a bath and a double bed 2. no WIFI only in the \\ lobby 3. room was to hot airco didn't worked properly \\ 4. really old fashion and this hotel urgently needs \\ to be refreshed 5. simple breakfast. \\ RESUMED this hotel does not deserve \\ 4 stars at all and can not be recommended at all. \\ We don't understand that booking.com included it in \\ its list. \textbf{Positive:} the location.\end{tabular} & 3.3  & 
\begin{tabular}[c]{@{}l@{}} 2.5, 3.3, 3.3, \\3.8, 4.2, 5.8, \\ 6.3 \end{tabular}\\ \hline
4 & \begin{tabular}[c]{@{}l@{}}\textbf{Negative:}. \textbf{Positive:} Location. Location. Location. \\ Room small but perfectly formed. \\ Staff very helpful and accommodated a change to \\ the offered menu. Decor modern and tasteful.\end{tabular} & 9.6  & 
\begin{tabular}[c]{@{}l@{}} 7.9, 8.8, 9.2, \\ 9.2, 9.6, 9.6, \\ 10 \end{tabular}\\ \hline
\end{tabular}
\caption{Example of four reviews, their scores, and the score distributions of all the reviews assigned to the same hotel. All the reviews are part of the train-validation data.}
 \label{review_examples}
\end{table}

\section {Equilibrium Analysis for Our Game}
\label{app:equ}

As promised in Section \ref{sec:verbal} we provide here the full details of the equilibria computations for our train and test sets.

We will now show the existence of an equilibrium in our setting, for both the train set hotels and the test set hotels (Table \ref{hotels_distributions}).  We will consider the decision maker's strategy $g(x)=choose$ iff $x > 8$ and 0 otherwise. 
The expert's strategy for the train set is defined as follows (in the notation below we refer to a situation by its expected value (EV)): 

\[f(EV) = \begin{cases}
 6.3 \ \mbox{if} \ EV = 4.17\\  
 8.3 \ \mbox{if} \ EV = 8.33\\ 
 9.2 \ \mbox{if} \ EV \in \{6.66, 7.97, 8.94, 9.19, 9.54\}\\ 
 9.6 \ \mbox{if} \ EV \in \{7.44, 8.11, 9.77\}
\end{cases}\]

To prove that $(f,g)$ is an equilibrium fix first $f$. We show that $g$ is a best-response. To see that observe that for every signal higher than 8 the average of all situations associated with it is higher than 8 (the cost). So, the only best response for that signal (i.e. for all situations associated with it, as the decision maker action depends only on the signal received) is {\em choose}. Obviously, in the only game associated with a received signal which is smaller than 8, the expected value is smaller than 8 and {\em skip} is a best-response.

Now, consider $g$ as fixed. We need to show that in no case the expert can lead to all 10 hotels be chosen. This will show the result as $f$ leads the decision maker to choose 9 hotels in $(f,g)$. 
Consider the hotel with an expected value of 4.17 and its possible signals. We need to show that no signal associated with this hotel leads to defining a set of situations (hotels) where the average expected values are as high as 8. This follows by case analysis: This hotel has only two signals in common with other hotels: 6.3 and 5.8; we will focus on these signals.

Consider the signal 6.3, if the expert chooses it as a recommendation only for hotel \#1, the average of the situation associated with it is 4.17, and hence the decision-maker best-response is {\em skip}. In another case, suppose the expert chooses this signal as a recommendation for two hotels, hotel \#1 and hotel \#2, hotel \#5 or hotel \#7, with an average score of 6.66, 8.11, or 8.95, respectively. In that case, the average of the situations associated with this signal is 5.41, 6.41, or 6.55. The expert can recommend this signal for hotel \#1 and two of the three hotels that shared this signal. In this case, the average of all the situations will be 6.31, 6.59, or 7.08 for all possible combinations. Finally, the expert can recommend the signal 6.3 for all the four hotels that shared it, and the average of these situations will be 6.97.

Consider the signal 5.8, if the expert chooses it as a recommendation only for hotel \#1, the average of the situation associated with it is 4.17, and hence the decision-maker best-response is {\em skip}. In another case, suppose the expert chooses this signal as a recommendation for two hotels, hotel \#1 and hotel \#3, hotel \#4 or hotel \#5, with an average score of 7.44, 7.97, or 8.11, respectively. In that case, the average of the situations associated with this signal is 5.8, 6.07, or 6.14. The expert can recommend this signal for hotel \#1 and two of the three hotels that shared this signal. In this case, the average of all the situations will be 6.52, 6.57, or 6.75 for all possible combinations. Finally, the expert can recommend the signal 6.3 for all the four hotels that shared it, and the average of these situations will be 6.92. Therefore, for any of the situation combinations, the average is still below 8, and the decision-maker's best response in these situations is {\em skip}. Thus, the expert payoff will be no more than 9.

As for the test set we consider the same $g$ as above and $f$ as defined below. The decision maker's strategy is the same as for the train set, i.e., $g(x)=choose$ iff $x > 8$ and 0 otherwise. The expert's strategy for the test set is as follows:
\[f(EV) = \begin{cases}
 6.3 \ \mbox{if} \ EV = 4.56\\  
 9.6 \ \mbox{if} \ EV \in \{8.04, 7.91, 8.33, 9.65\} \\
 10 \ \mbox{if} \ EV \in \{6.69, 7.74, 8.89, 9.04, 9.66\}\\ 
\end{cases}\]

As before, to prove that $(f,g)$ is an equilibrium we fix first $f$ and show that $g$ is a best-response.  Observe that for every signal higher than 8 the average of all situations associated with it is higher than 8 and the only best response for that signal is {\em choose}. In the only game associated with a received signal which is smaller than 8 the expected value is smaller than 8 and {\em skip} is a best-response.

Now, consider $g$ as fixed. We need to show that in no case the expert can lead to all 10 hotels be chosen.
A similar case analysis as done before can be done for the test set. Consider the hotel with an expected value of 4.56 and its possible signals. We need to show that no signal associated with this hotel leads to defining a set of situations (hotels) where the average expected values are as high as 8. This follows by case analysis: This hotel has only two signals in common with other hotels: 6.3 and 4.2; we will focus on these signals.

Consider the signal 6.3, if the expert chooses it as a recommendation only for hotel \#1, the average of the situation associated with it is 4.56, and hence the decision-maker best-response is {\em skip}. In another case, suppose the expert chooses this signal as a recommendation for two hotels, hotel \#1 and hotel \#2 or hotel \#7, with an average score of 6.69, or 8.89, respectively. In that case, the average of the situations associated with this signal is 5.62, or 6.72. Finally, the expert can recommend the signal 6.3 for all the three hotels that shared it, and the average of these situations will be 6.71.

Consider the signal 4.2, if the expert chooses it as a recommendation only for hotel \#1, the average of the situation associated with it is 4.56, and hence the decision-maker's best-response is {\em skip}. Finally, the expert can choose this signal for two hotels that share it, i.e., hotel \#1 and hotel \#2, with average scores of 4.56 and 6.69. This combination of situations will result in an average of 5.62. Therefore, for any of the situation combinations, the average is still below 8, and the decision-maker's best response in these situations is {\em skip}. Thus, the expert payoff will be no more than 9.

\section{Features}
\label{appendix:features}

\begin{table}[H]
\centering
\begin{tabular}{|l|l|l|l|}
\hline & \textbf{Group \#1} & \textbf{Group \#2} & \textbf{Group \#3} \\ \hline
\textbf{\begin{tabular}[c]{@{}l@{}}Positive\\ Part\end{tabular}} & \begin{tabular}[c]{@{}l@{}}comfort, ok, close, like, \\sleek, spacious, available, \\ fair, welcoming, helpful,\\ new, lots, extra, neat,\\ tastefully, cleanliness, \\ friendly, good, variety, \\ quiet, approachable,\\ refurbished, surprising, \\ free, honest, attentive,\\ comfy, comfortable, \\ newly, clean, nicely \\ well, easy, big\\ plenty, safe, nice\end{tabular} & \begin{tabular}[c]{@{}l@{}}very, large, surprise, \\ great, really, love,\\ renovated, huge,\\ unrivalled, pristine,\\ enjoyed, impressive,\\ extensive, warm, \\ brilliant, exceeded \\ professional, useful, \\ efficiently, highly \\ everything\end{tabular} & \begin{tabular}[c]{@{}l@{}}stunning, superb, \\ extremely, impressed, \\ fabulous, perfectly,\\ perfect, awesome, \\ outstanding, wow, \\ incredible, top class, \\ beyond fabulous, best, \\ phenomenal, special, \\ spectacular, fab, \\ magnificent, fantastic, \\ largest, spotless,\\ excellent, amazing\end{tabular} \\ \hline
\textbf{\begin{tabular}[c]{@{}l@{}}Negative\\ Part\end{tabular}} & \begin{tabular}[c]{@{}l@{}}little, small, overpriced,\\ could be better, dirty,\\ supposed, tacky,\\ not friendly, basic,\\ outdated, bit of,\\ noisy, expensive, simple,\\limited, smell, old,\\ tight, missing, compact,\\inconspicuous, busy,\\ isolated, average,\\ pricier, noise, thin, not\\ obsolete, sleepy, empty, \\ uncomfortable, regular
\end{tabular} & \begin{tabular}[c]{@{}l@{}}very, everything,\\ not good enough,\\ too, bad, really, \\ poor, sufficiently\\ claustrophobic\end{tabular} & \begin{tabular}[c]{@{}l@{}}incredibly, shockingly, \\ overcharged, awful, \\ disgusting, \\ worst, completely, \\ flooding, extremely, \\ surprised, dangerous, \\ exceptional, urgently, \\ terrible, disaster,\\ disappointed\end{tabular} \\ \hline
\end{tabular}
\caption{Sentiment word groups used in the hand-crafted features.}
\label{tab:semantic_group}
\end{table}

\begin{center}
\begin{longtable}[]{|c|c|c|c|}
\hline
\textbf{\begin{tabular}[c]{@{}c@{}}High\\ Level\end{tabular}} & \textbf{\#} & \textbf{\begin{tabular}[c]{@{}c@{}}Feature\\ Name\end{tabular}} & \textbf{\begin{tabular}[c]{@{}c@{}}Feature\\ Description\end{tabular}} \\ \hline
& 1 & Facilities & \begin{tabular}[c]{@{}c@{}}The positive part provides details \\ about the hotel's facilities\end{tabular} \\ 
\cline{2-4} & 2 & Price & \begin{tabular}[c]{@{}c@{}}The positive part provides details \\ about the hotel's price\end{tabular}  \\ 
\cline{2-4} \multirow{3}{*}{\textbf{\begin{tabular}[c]{@{}c@{}}Positive \\ Part\\ Topics\end{tabular}}} & 3 & Design & \begin{tabular}[c]{@{}c@{}}The positive part provides details\\  about the hotel's design\end{tabular}\\ 
\cline{2-4} & 4 & Location & \begin{tabular}[c]{@{}c@{}}The positive part provides details\\  about the hotel's location\end{tabular} \\ 
\cline{2-4} & 5 & Room & \begin{tabular}[c]{@{}c@{}}The positive part provides details\\  about the room\end{tabular} \\ 
\cline{2-4} & 6 & Staff & \begin{tabular}[c]{@{}c@{}}The positive part provides details \\ about the hotel's staff\end{tabular} \\ 
\cline{2-4} & 7 & Food & \begin{tabular}[c]{@{}c@{}}The positive part provides details \\ about the food in the hotel\end{tabular} \\ 
\cline{2-4} & 8 & Transportation & \begin{tabular}[c]{@{}c@{}}The positive part provides details \\ about the transportation \\ options in the hotel's area\end{tabular} \\ 
\cline{2-4} & 9 & \begin{tabular}[c]{@{}c@{}}Sanitary\\ Facilities\end{tabular} & \begin{tabular}[c]{@{}c@{}}The positive part provides \\ details  about the \\ sanitary facilities in the room\end{tabular} \\ 
\cline{2-4} & 10 & View & \begin{tabular}[c]{@{}c@{}}The positive part provides details \\ about the view from the hotel\end{tabular}\\ 
\hline & 11 & Empty & The positive part is empty \\ 
\cline{2-4} & 12 & Nothing positive & \begin{tabular}[c]{@{}c@{}}The positive part\\explicitly states that there is \\ nothing positive about the hotel\end{tabular} \\ 
\cline{2-4} & 13 & Summary Sentence & \begin{tabular}[c]{@{}c@{}}The positive part provides \\ a positive summary sentence, \\ e.g., “We would stay there again”\end{tabular} \\ 
\cline{2-4} & 14 & \begin{tabular}[c]{@{}c@{}}Words from the\\first positive group\end{tabular} & \begin{tabular}[c]{@{}c@{}}The positive part contains\\ words from the first positive group\end{tabular} \\ 
\cline{2-4} & 15 & \begin{tabular}[c]{@{}c@{}}Words from the\\second positive group\end{tabular} & \begin{tabular}[c]{@{}c@{}}The positive part contains\\  words from the second positive group\end{tabular} \\ 
\cline{2-4} \multirow{-8}{*}{\textbf{\begin{tabular}[c]{@{}c@{}}Positive \\ Part\\ Properties\end{tabular}}} & 16 & \begin{tabular}[c]{@{}c@{}}Words from the\\ third positive group\end{tabular} & \begin{tabular}[c]{@{}c@{}}The positive part contains\\  words from the third positive group\end{tabular} \\ 
\cline{2-4} & 17 & Short positive part & \begin{tabular}[c]{@{}c@{}}The number of characters \\ in the positive part is lower than 100\end{tabular} \\ 
\cline{2-4} & 18 & Medium positive part & \begin{tabular}[c]{@{}c@{}}The number of characters in the \\ positive part is between 100 and 199\end{tabular} \\ 
\cline{2-4} & 19 & Long positive part & \begin{tabular}[c]{@{}c@{}}The number of characters in the \\ positive part is higher than 200\end{tabular} \\ 
\hline & 20 & Price & \begin{tabular}[c]{@{}c@{}}The negative part provides\\details about the hotel's price\end{tabular} \\ 
\cline{2-4} & 21 & Staff & \begin{tabular}[c]{@{}c@{}}The negative part provides\\details about the hotel's staff\end{tabular} \\ 
\cline{2-4} \multirow{4}{*}{\textbf{\begin{tabular}[c]{@{}c@{}}Negative\\ Part\\ Topics\end{tabular}}} & 22 & \begin{tabular}[c]{@{}c@{}}Sanitary\\ Facilities\end{tabular} & \begin{tabular}[c]{@{}c@{}}The negative part \\ provides details about the \\ sanitary facilities in the room\end{tabular} \\ 
\cline{2-4} & 23 & Room & \begin{tabular}[c]{@{}c@{}}The negative part provides\\details about the room\end{tabular} \\ 
\cline{2-4} & 24 & Food & \begin{tabular}[c]{@{}c@{}}The negative part provides\\details about the food in the hotel\end{tabular} \\ 
\cline{2-4} & 25 & Location & \begin{tabular}[c]{@{}c@{}}The negative part provides\\details about the hotel's location\end{tabular} \\ 
\cline{2-4} & 26 & Facilities & \begin{tabular}[c]{@{}c@{}}The negative part provides\\details about the hotel's facilities\end{tabular} \\ 
\cline{2-4} & 27 & Air & \begin{tabular}[c]{@{}c@{}}The negative part\\provides details about the \\ hotel's air-conditioning facilities\end{tabular} \\ 
\hline & 28 & Empty & The negative part is empty \\ 
\cline{2-4} & 29 & Nothing negative & \begin{tabular}[c]{@{}c@{}}The negative part \\explicitly states that there is \\ nothing negative about the hotel\end{tabular} \\ 
\cline{2-4} & 30 & Summary Sentence & \begin{tabular}[c]{@{}c@{}}The negative part provides \\ a negative summary sentence, \\ e.g., “I do not know \\ how it is a 4 stars hotel”\end{tabular} \\ 
\cline{2-4} & 31 & \begin{tabular}[c]{@{}c@{}}Words from the\\ first negative group\end{tabular} & \begin{tabular}[c]{@{}c@{}}The negative part provides\\ words from the first negative group\end{tabular} \\ 
\cline{2-4} & 32 & \begin{tabular}[c]{@{}c@{}}Words from the\\ second negative group\end{tabular} & \begin{tabular}[c]{@{}c@{}}The negative part provides \\ words from the second negative group\end{tabular} \\ 
\cline{2-4} \multirow{-8}{*}{\textbf{\begin{tabular}[c]{@{}c@{}}Negative\\ Part\\ Properties\end{tabular}}} & 33 & \begin{tabular}[c]{@{}c@{}}Words from the\\ third negative group\end{tabular} & \begin{tabular}[c]{@{}c@{}}The negative part provides \\ words from the third negative group\end{tabular} \\ 
\cline{2-4} & 34 & Short negative part & \begin{tabular}[c]{@{}c@{}}The number of characters in the \\ negative part is lower than 100\end{tabular} \\ 
\cline{2-4} & 35 & Medium negative part & \begin{tabular}[c]{@{}c@{}}The number of characters in the \\ negative part is between 100 and 199\end{tabular} \\ 
\cline{2-4} & 36 & Long negative part & \begin{tabular}[c]{@{}c@{}}The number of characters in the \\ negative part is higher than 200\end{tabular} \\ 
\hline {\color[HTML]{333333}} & 37 & \begin{tabular}[c]{@{}c@{}}Detailed Review\end{tabular} & \begin{tabular}[c]{@{}c@{}}The review provides many\\ details about the hotel\end{tabular} \\ 
\cline{2-4} \multirow{-1}{*}{{\color[HTML]{333333} \textbf{\begin{tabular}[c]{@{}c@{}}Overall\\ Review\\ Properties\end{tabular}}}} {\color[HTML]{333333}} & 38 & \begin{tabular}[c]{@{}c@{}}Review structured\\ as a list\end{tabular} & \begin{tabular}[c]{@{}c@{}}The review is arranged as a list of the \\ hotel's positive and negative properties\end{tabular} \\ 
\cline{2-4} {\color[HTML]{333333}} & 39 & \begin{tabular}[c]{@{}c@{}}Positive part \\ shown first\end{tabular} & \begin{tabular}[c]{@{}c@{}}The positive part is shown \\ before the negative part\end{tabular} \\ 
\cline{2-4} & 40 & \begin{tabular}[c]{@{}c@{}}Low proportion\\between positive and\\negative parts' lengths\end{tabular}  & \begin{tabular}[c]{@{}c@{}}The proportion between the number \\ of characters in the positive and \\ the negative parts is lower than 0.7\end{tabular} \\ 
\cline{2-4} {\color[HTML]{333333}} & 41 & \begin{tabular}[c]{@{}c@{}}Medium proportion\\between positive and\\negative parts' lengths\end{tabular} & \begin{tabular}[c]{@{}c@{}}The proportion between the number \\ of characters in the positive and \\ the negative parts is between 0.7 and 4 \end{tabular} \\ 
\cline{2-4} {\color[HTML]{333333}} & 42 & \begin{tabular}[c]{@{}c@{}}High proportion\\between positive and\\negative parts' lengths\end{tabular} & \begin{tabular}[c]{@{}c@{}}The proportion between the number \\ of characters in the positive and \\ the negative parts is higher than 4
\end{tabular} \\ 
\hline
\caption{Descriptions of the hand-crafted features.}
  \label{tab:Domain_based_attributes_description}
\end{longtable}
\end{center}

\begin{center}
\begin{longtable}{|c|c|c|c|c|c|c|}
\hline
\textbf{\begin{tabular}[c]{@{}c@{}}High\\ Level\end{tabular}} & \textbf{\#} & \textbf{\begin{tabular}[c]{@{}c@{}}Feature Name\end{tabular}} & \textbf{\begin{tabular}[c]{@{}c@{}}Text \\ \#1\end{tabular}} & \textbf{\begin{tabular}[c]{@{}c@{}}Text \\ \#2\end{tabular}} & \textbf{\begin{tabular}[c]{@{}c@{}}Text\\  \#3\end{tabular}} & \textbf{\begin{tabular}[c]{@{}c@{}}Text \\ \#4\end{tabular}} \\ \hline
& 1 & Facilities & 0 & 1 & 0 & 0 \\ 
\cline{2-7} & 2 & Price & 0 & 0 & 0 & 0 \\ 
\cline{2-7} & 3 & Design & 0 & 0 & 0 & 1 \\ 
\cline{2-7} & 4 & Location & 1 & 0 & 1 & 1 \\ 
\cline{2-7} & 5 & Room & 0 & 1 & 0 & 1 \\ 
\cline{2-7} & 6 & Staff & 0 & 0 & 0 & 1 \\ 
\cline{2-7} & 7 & Food & 0 & 0 & 0 & 0 \\ 
\cline{2-7} & 8 & Transportation & 0 & 1 & 0 & 0 \\ 
\cline{2-7} & 9 & Sanitary Facilities & 0 & 1 & 0 & 0 \\ 
\cline{2-7} \multirow{-10}{*}{\textbf{\begin{tabular}[c]{@{}c@{}}Positive \\ Part\\ Topics\end{tabular}}} 
& 10 & View & 0 & 1 & 0 & 0 \\ 
\hline & 11 & Empty & 0 & 0 & 0 & 0 \\ 
\cline{2-7} & 12 & Nothing positive & 0 & 0 & 0 & 0 \\ 
\cline{2-7} & 13 & Summary Sentence & 0 & 1 & 0 & 0 \\ 
\cline{2-7} & 14 & \begin{tabular}[c]{@{}c@{}} Words from the first\\positive group \end{tabular} & 0 & 1 & 0 & 1 \\ 
\cline{2-7} & 15 & \begin{tabular}[c]{@{}c@{}} Words from the second\\positive group \end{tabular} & 0 & 1 & 0 & 1 \\ 
\cline{2-7} & 16 & \begin{tabular}[c]{@{}c@{}} Words from the third\\positive group \end{tabular} & 1 & 1 & 0 & 1 \\ 
\cline{2-7} \multirow{-8}{*}{\textbf{\begin{tabular}[c]{@{}c@{}}Positive \\ Part\\ Properties\end{tabular}}}  & 17 & Short positive part & 0 & 1 & 0 & 0 \\ 
\cline{2-7} & 18 & Medium positive part & 0 & 0 & 1 & 0 \\ 
\cline{2-7} & 19 & Long positive part & 1 & 0 & 0 & 1 \\ 
\hline & 20 & Price & 0 & 0 & 0 & 0 \\ 
\cline{2-7} & 21 & Staff & 1 & 0 & 0 & 0 \\
\cline{2-7} & 22 & Sanitary Facilities & 0 & 0 & 0 & 0 \\
\cline{2-7} \multirow{-1}{*}{\textbf{\begin{tabular}[c]{@{}c@{}}Negative\\ Part\\ Topics\end{tabular}}} & 23 & Room & 0 & 0 & 1 & 0 \\ 
\cline{2-7} & 24 & Food & 0 & 0 & 1 & 0 \\ 
\cline{2-7} & 25 & Location & 0 & 0 & 0 & 0 \\
\cline{2-7} & 26 & Facilities & 1 & 0 & 0 & 0 \\
\cline{2-7} & 27 & Air & 0 & 0 & 1 & 0 \\ 
\hline & 28 & Empty & 0 & 0 & 0 & 1 \\ 
\cline{2-7} & 29 & Nothing negative & 0 & 1 & 0 & 0 \\ 
\cline{2-7} & 30 & Summary Sentence & 0 & 0 & 1 & 0 \\ 
\cline{2-7} & 31 & \begin{tabular}[c]{@{}c@{}} Words from the first\\negative group \end{tabular} & 1 & 1 & 1 & 0 \\ 
\cline{2-7} & 32 & \begin{tabular}[c]{@{}c@{}} Words from the second\\negative group \end{tabular} & 1 & 1 & 1 & 0 \\ 
\cline{2-7} & 33 &\begin{tabular}[c]{@{}c@{}} Words from the third\\negative group \end{tabular} & 1 & 0 & 1 & 0 \\ 
\cline{2-7} \multirow{-8}{*}{\textbf{\begin{tabular}[c]{@{}c@{}}Negative\\ Part\\ Properties\end{tabular}}} & 34 & Short negative part & 0 & 0 & 0 & 0 \\ 
\cline{2-7} & 35 & Medium negative part & 0 & 1 & 0 & 1 \\ 
\cline{2-7} & 36 & Long negative part & 1 & 0 & 1 & 0 \\ 
\hline & 37 & Detailed Review & 1 & 1 & 0 & 0 \\ 
\cline{2-7} & 38 & Review structured as a list & 0 & 0 & 1 & 0 \\ 
\cline{2-7} & 39 & Positive part shown first & 0 & 1 & 0 & 0 \\  
\cline{2-7} 
\multirow{-2}{*}{{\textbf{\begin{tabular}[c]{@{}c@{}}Overall\\ Review\\ Properties\end{tabular}}}}
& 40 & \begin{tabular}[c]{@{}c@{}}Low proportion\\between positive and\\negative parts' lengths\end{tabular}  & 0 & 1 & 0 & 1 \\ 
\cline{2-7} & 41 & \begin{tabular}[c]{@{}c@{}}Medium proportion\\between positive and\\negative parts' lengths\end{tabular} & 1 & 0 & 1 & 0 \\
\cline{2-7} & 42 & \begin{tabular}[c]{@{}c@{}}High proportion\\between positive and\\negative parts' lengths\end{tabular} & 0 & 0 & 0 & 0 \\ \hline

\caption{Hand-crafted feature values for the texts of Appendix \ref{appendix:data}.}
  \label{tab:Domain_based_attributes}
\end{longtable}
\end{center}

\bibliographystyle{theapa}
\bibliography{inputs/bibdb}

\end{document}

%% file: inputs/defs.tex
\usepackage{xcolor}
\usepackage{gensymb} 
\usepackage{graphicx}
\usepackage{amsmath}
\usepackage{dsfont}
\usepackage{multirow} 
\usepackage{amssymb} 

\newcommand{\mA}{\mathcal{A}}
\newcommand{\mB}{\mathcal{B}}

\newcommand{\mT}{\mathcal{T}}

\newcommand{\ind}{\mathds{1}}

\newcommand{\hr}{\mathcal{HR}}

\newcommand{\ts}{\mathcal{TS}}
\newcommand{\hs}{\mathcal{HS}}

\citepunct{}{\&}{and}{, }{; }{, }{}{}{.}
\newcommand{\citep}[1]{(\cite{#1})}
\newcommand{\citet}[1]{\citeauthor{#1}~(\citeyear{#1})}

\numberwithin{claim}{section} 
\numberwithin{observation}{section} 
\numberwithin{question}{section}

\DeclareSymbolFont{rsfs}{U}{rsfs}{m}{n}
\DeclareSymbolFontAlphabet{\mathscrsfs}{rsfs}
\DeclareUnicodeCharacter{2212}{-}